\documentclass[final,3p,authoryear,times,twocolumn]{elsarticle}
\usepackage{graphicx}
\usepackage{url}
\usepackage{multirow}
\usepackage{booktabs}
\usepackage{graphicx,epsfig,subfig}
\usepackage{amsmath,amssymb,amsfonts,amsthm}
\usepackage{bbm}
\usepackage{todonotes}
\usepackage{longtable}
\usepackage{dblfloatfix}
\usepackage{xcolor}
\usepackage[colorlinks=true,linkcolor=blue,citecolor=blue,urlcolor=blue,pagebackref=false]{hyperref}
\begin{document}

\begin{frontmatter}
\title{A Survey on Deep Learning in Medical Image Analysis}

\author{Geert Litjens, Thijs Kooi, Babak Ehteshami Bejnordi, Arnaud Arindra Adiyoso Setio, Francesco Ciompi,\\Mohsen Ghafoorian, Jeroen A.W.M. van der Laak, Bram van Ginneken, Clara I. S\'{a}nchez\\
\hspace*{1cm}\\
Diagnostic Image Analysis Group\\
Radboud University Medical Center\\
Nijmegen, The Netherlands
}

\begin{abstract}
Deep learning algorithms, in particular convolutional networks, have rapidly become a methodology of choice for analyzing medical images. This paper reviews the major deep learning concepts pertinent to medical image analysis and summarizes over 300 contributions to the field, most of which appeared in the last year. We survey the use of deep learning for image classification, object detection, segmentation, registration, and other tasks. Concise overviews are provided of studies per application area: neuro, retinal, pulmonary, digital pathology, breast, cardiac, abdominal, musculoskeletal. We end with a summary of the current state-of-the-art, a critical discussion of open challenges and directions for future research.
\end{abstract}

\end{frontmatter}

\section{Introduction}
As soon as it was possible to scan and load medical images into a computer, researchers have built systems for automated analysis. Initially, from the 1970s to the 1990s, medical image analysis was done with sequential application of low-level pixel processing (edge and line detector filters, region growing) and mathematical modeling (fitting lines, circles and ellipses) to construct compound rule-based systems that solved particular tasks. There is an analogy with expert systems with many if-then-else statements that were popular in artificial intelligence in the same period. These expert systems have been described as GOFAI (good old-fashioned artificial intelligence) \citep{Haug85} and were often brittle; similar to rule-based image processing systems.  

At the end of the 1990s, supervised techniques, where training data is used to develop a system, were becoming increasingly popular in medical image analysis. Examples include active shape models (for segmentation), atlas methods (where the atlases that are fit to new data form the training data), and the concept of feature extraction and use of statistical classifiers (for computer-aided detection and diagnosis). This pattern recognition or machine learning approach is still very popular and forms the basis of many successful commercially available medical image analysis systems. Thus, we have seen a shift from systems that are completely designed by humans to systems that are trained by computers using example data from which feature vectors are extracted. Computer algorithms determine the optimal decision boundary in the high-dimensional feature space. A crucial step in the design of such systems is the extraction of discriminant features from the images. This process is still done by human researchers and, as such, one speaks of systems with {\em handcrafted} features. 

A logical next step is to let computers learn the features that optimally represent the data for the problem at hand. This concept lies at the basis of many deep learning algorithms: models (networks) composed of many layers that transform input data (e.g.\ images) to outputs (e.g.\ disease present/absent) while learning increasingly higher level features. The most successful type of models for image analysis to date are convolutional neural networks (CNNs). CNNs contain many layers that transform their input with convolution filters of a small extent. Work on CNNs has been done since the late seventies \citep{Fuku80} and they were already applied to medical image analysis in 1995 by \cite{Lo95}. They saw their first successful real-world application in LeNet \citep{Lecu98} for hand-written digit recognition. Despite these initial successes, the use of CNNs did not gather momentum until various new techniques were developed for efficiently training deep networks, and advances were made in  core computing systems. The watershed was the contribution of \cite{Kriz12} to the ImageNet challenge in December 2012. The proposed CNN, called AlexNet, won that competition by a large margin. In subsequent years, further progress has been made using related but deeper architectures \citep{Russ14a}. In computer vision, deep convolutional networks have now become the technique of choice.

The medical image analysis community has taken notice of these pivotal developments. However, the transition from systems that use handcrafted features to systems that learn features from the data has been gradual. Before the breakthrough of AlexNet, many different techniques to learn features were popular. \cite{Beng13} provide a thorough review of these techniques. They include principal component analysis, clustering of image patches, dictionary approaches, and many more. \cite{Beng13} introduce CNNs that are trained end-to-end only at the end of their review in a section entitled \textit{Global training of deep models}. In this survey, we focus particularly on such deep models, and do not include the more traditional feature learning approaches that have been applied to medical images. For a broader review on the application of deep learning in health informatics we refer to \cite{Ravi17}, where medical image analysis is briefly touched upon.

\begin{figure*}[!tb]
\centering
\includegraphics[width=0.49\textwidth]{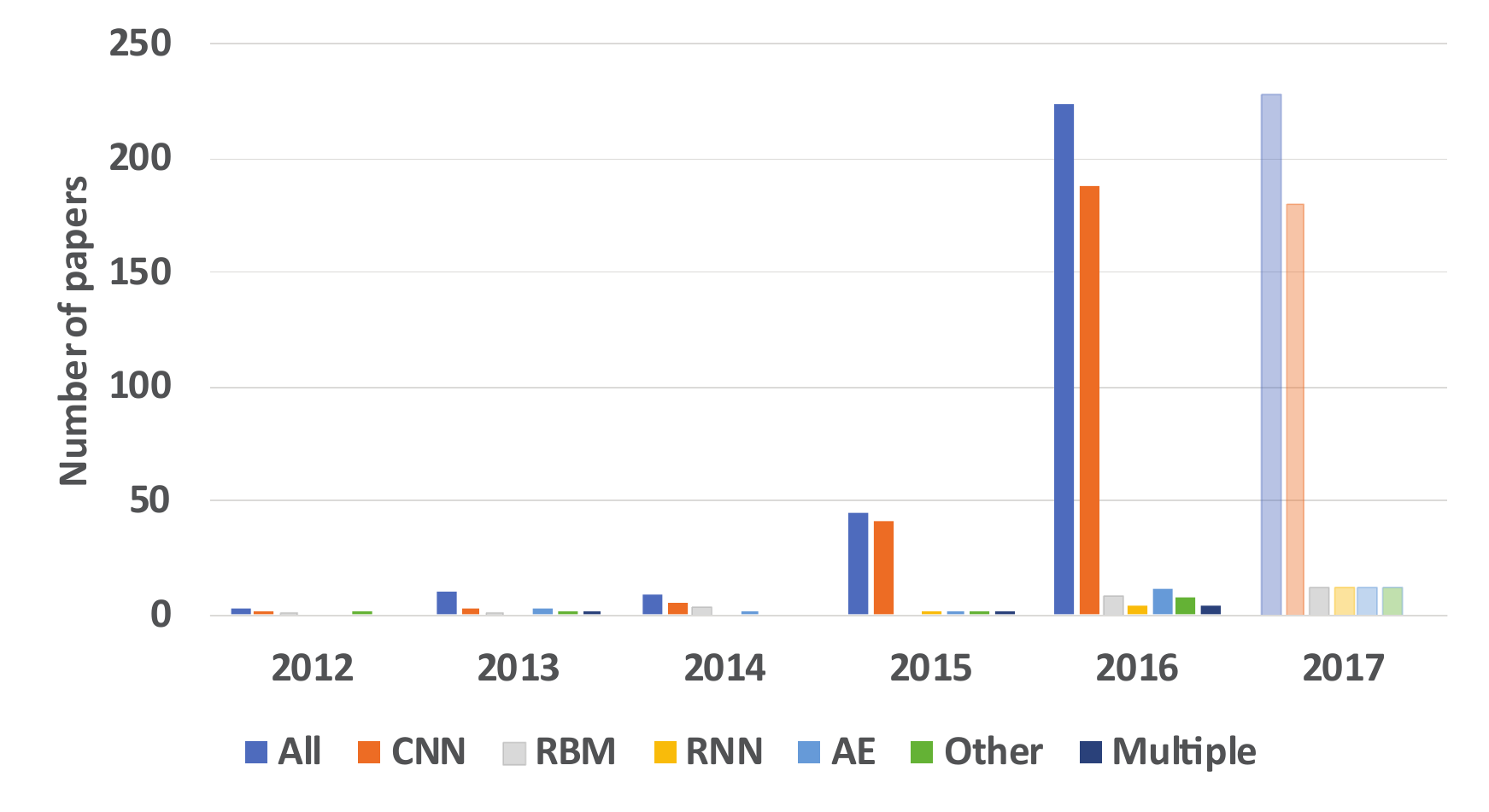}
\includegraphics[width=0.49\textwidth]{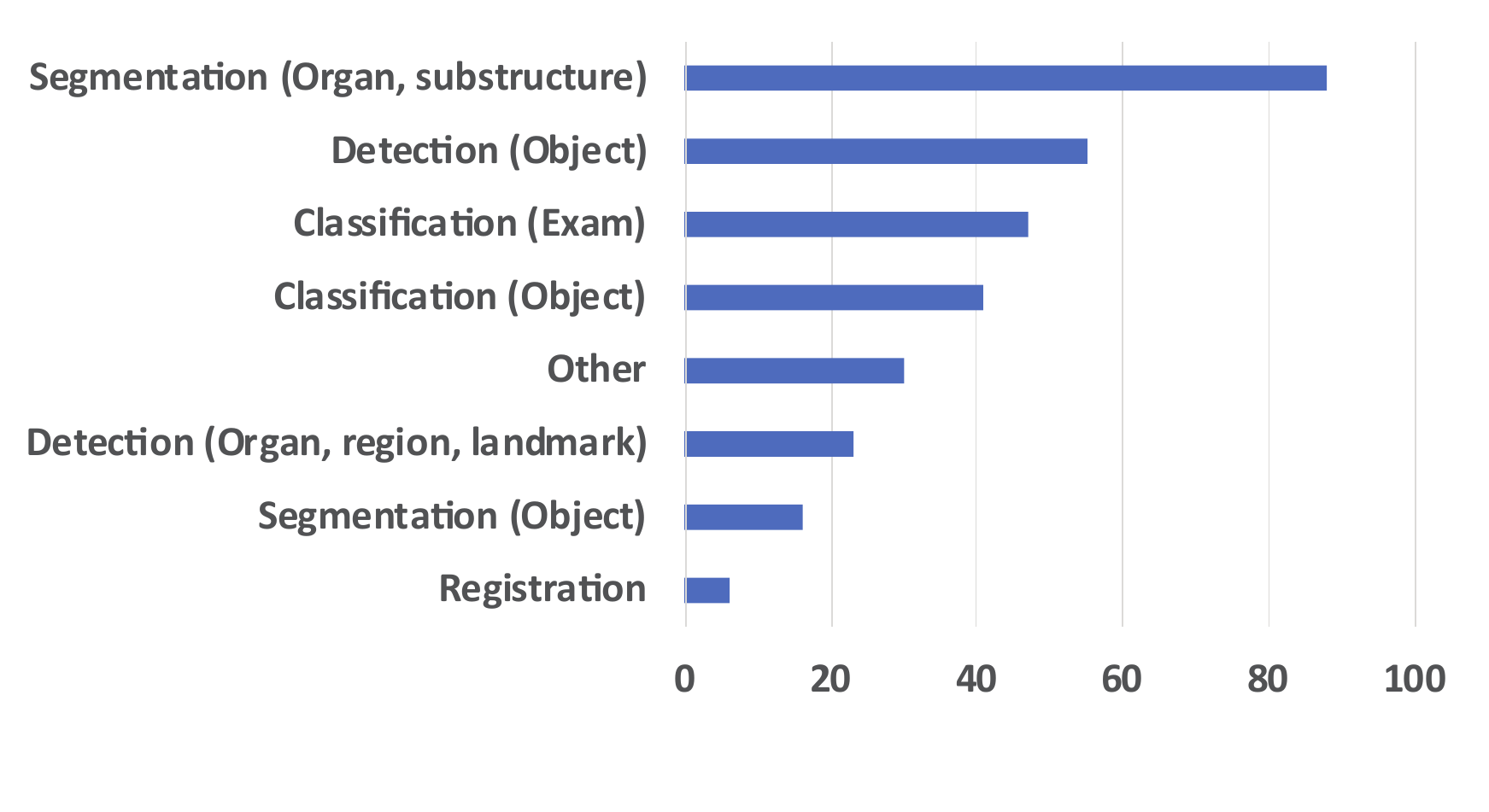}
\includegraphics[width=0.49\textwidth]{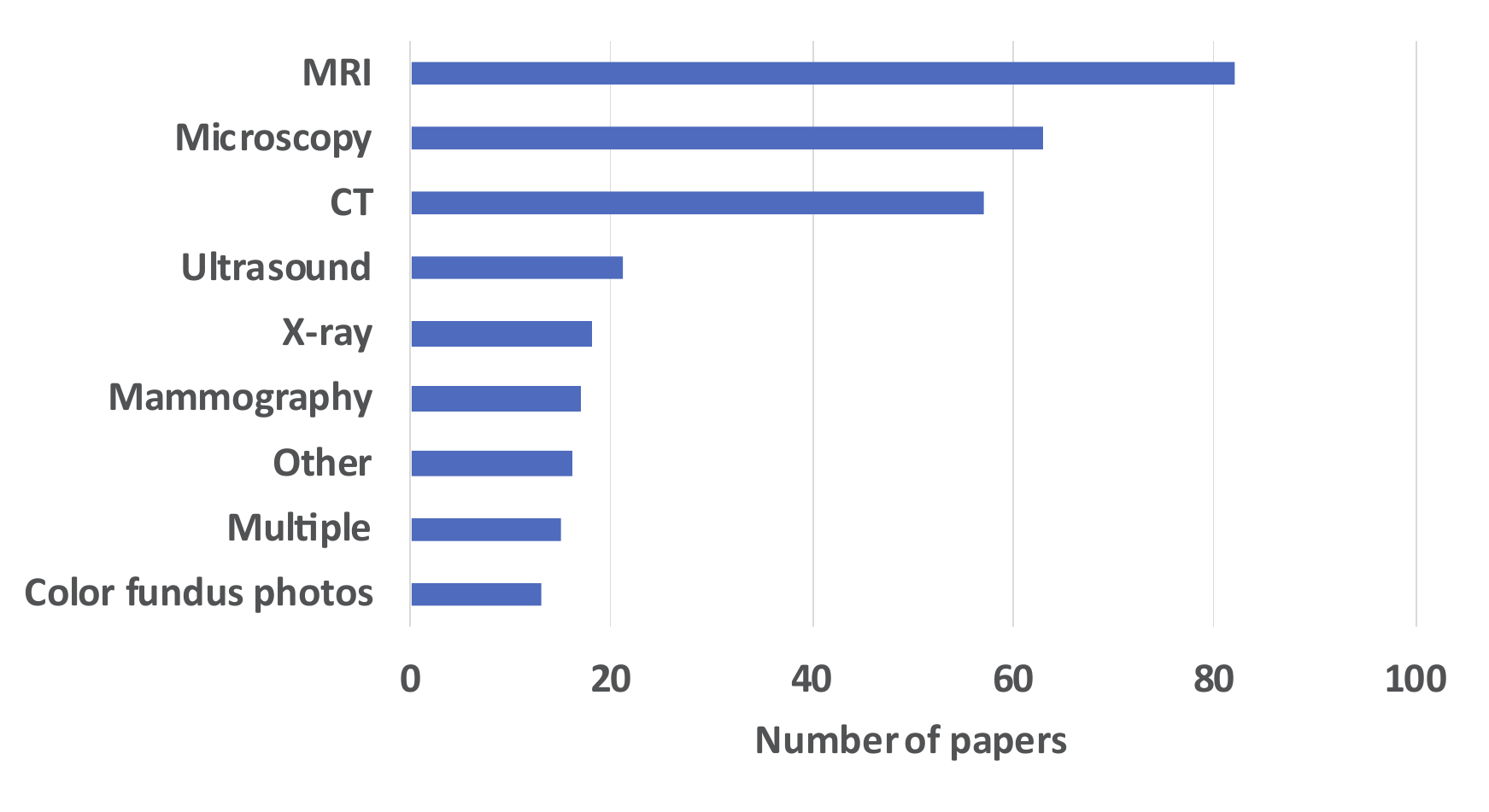}
\includegraphics[width=0.49\textwidth]{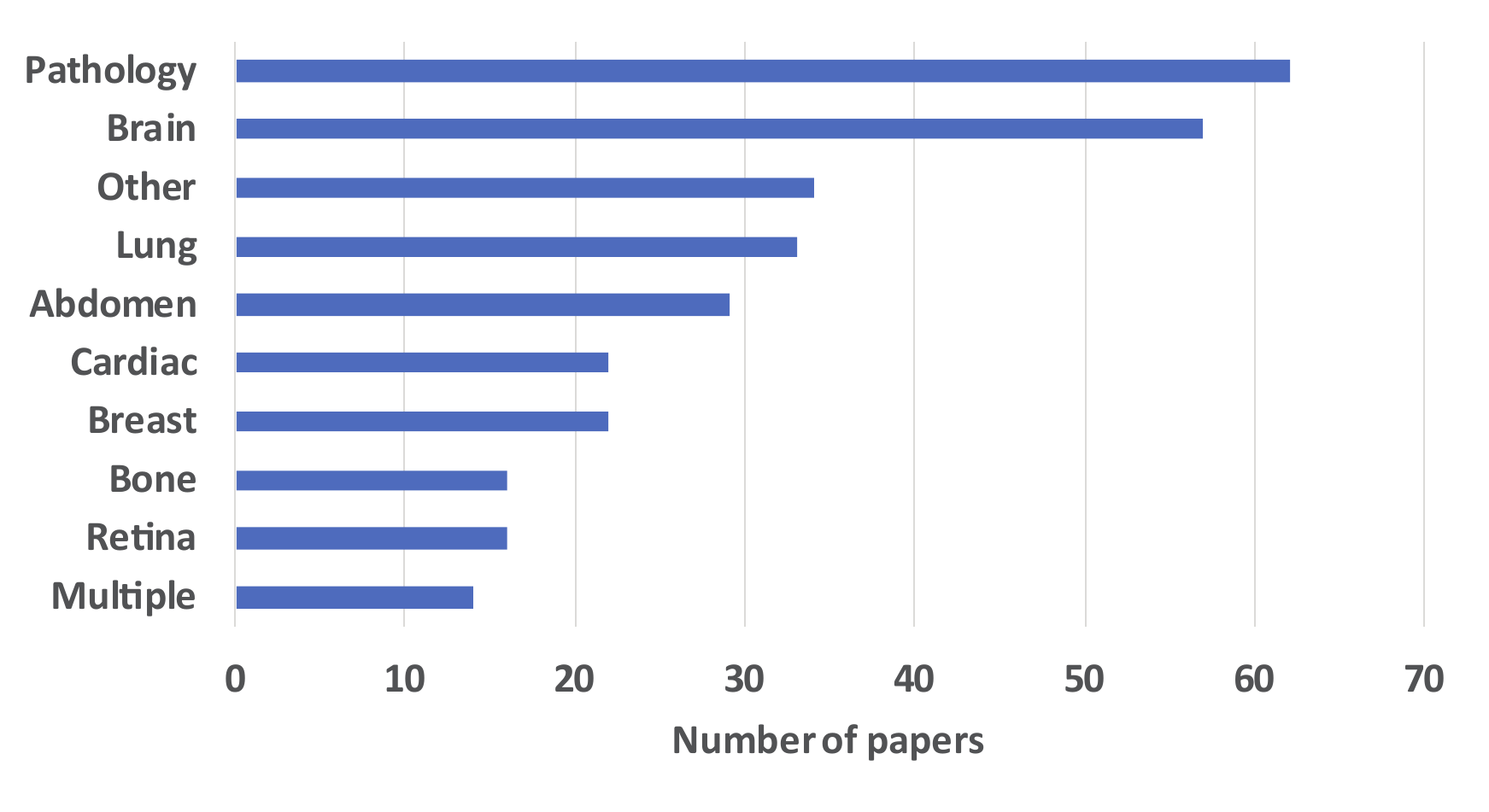}
\caption{Breakdown of the papers included in this survey in year of publication, task addressed (Section \ref{dl-mi}), imaging modality, and application area (Section \ref{sec:app}). The number of papers for 2017 has been extrapolated from the papers published in January.
\label{fig:statistics}}
\end{figure*}

Applications of deep learning to medical image analysis first started to appear at workshops and conferences, and then in journals. The number of papers grew rapidly in 2015 and 2016. This is illustrated in Figure \ref{fig:statistics}. The topic is now dominant at major conferences and a first special issue appeared of IEEE Transaction on Medical Imaging in May 2016 \citep{Gree16}.

\color{black}
One dedicated review on application of deep learning to medical image analysis was published by \cite{Shen17}. Although they cover a substantial amount of work, we feel that important areas of the field were not represented. To give an example, no work on retinal image analysis was covered. The motivation for our review was to offer a comprehensive overview of (almost) all fields in medical imaging, both from an application and a methodology-drive perspective. This also includes overview tables of all publications which readers can use to quickly assess the field. Last, we leveraged our own experience with the application of deep learning methods to medical image analysis to provide readers with a dedicated discussion section covering the state-of-the-art, open challenges and overview of research directions and technologies that will become important in the future.
\color{black}

\color{black}
This survey includes over 300 papers, most of them recent, on a wide variety of applications of deep learning in medical image analysis. To identify relevant contributions PubMed was queried for papers containing ("convolutional" OR "deep learning") in title or abstract. ArXiv was searched for papers mentioning one of a set of terms related to medical imaging. Additionally, conference proceedings for MICCAI (including workshops), SPIE, ISBI and EMBC were searched based on titles of papers. We checked references in all selected papers and consulted colleagues. We excluded papers that did not report results on medical image data or only used standard feed-forward neural networks with handcrafted features. When overlapping work had been reported in multiple publications, only the publication(s) deemed most important were included. We expect the search terms used to cover most, if not all, of the work incorporating deep learning methods. The last update to the included papers was on February 1, 2017. The appendix describes the search process in more detail. 
\color{black}

Summarizing, with this survey we aim to: 
\begin{itemize}
\item show that deep learning techniques have permeated the entire field of medical image analysis;
\item identify the challenges for successful application of deep learning to medical imaging tasks;
\item highlight specific contributions which solve or circumvent these challenges.
\end{itemize}

The rest of this survey as structured as followed. In Section \ref{sec:neuralnets} we introduce the main deep learning techniques that have been used for medical image analysis and that are referred to throughout the survey. Section \ref{dl-mi} describes the contributions of deep learning to canonical tasks in medical image analysis: classification, detection, segmentation, registration, retrieval, image generation and enhancement. Section \ref{sec:app} discusses obtained results and open challenges in different application areas: neuro, ophthalmic, pulmonary, digital pathology and cell imaging, breast, cardiac, abdominal, musculoskeletal, and remaining miscellaneous applications. We end with a summary, a critical discussion and an outlook for future research.

\color{black}
\section{Overview of deep learning methods}
\label{sec:neuralnets}
The goal of this section is to provide a formal introduction and definition of the deep learning concepts, techniques and architectures that we found in the medical image analysis papers surveyed in this work. 

\subsection{Learning algorithms}
Machine learning methods are generally divided into {\it supervised} and {\it unsupervised learning} algorithms, although there are many nuances. In supervised learning, a model is presented with a dataset $\mathcal{D} = \{{\bf x}, y\}^{N}_{n = 1}$ of input features ${\bf x}$ and label $y$ pairs, where $y$ typically represents an instance of a fixed set of classes. In the case of regression tasks $y$ can also be a vector with continuous values. Supervised training typically amounts to finding model parameters $\Theta$ that best predict the data based on a loss function $L(y, \hat{y})$. Here $\hat{y}$ denotes the output of the model obtained by feeding a data point ${\bf x}$ to the function $f({\bf x}; \Theta)$ that represents the model.

Unsupervised learning algorithms process data without labels and are trained to find patterns, such as latent subspaces. Examples of traditional unsupervised learning algorithms are principal component analysis and clustering methods. Unsupervised training can be performed under many different loss functions. One example is reconstruction loss $L({\bf x}, \hat{{\bf x}})$ where the model has to learn to reconstruct its input, often through a lower-dimensional or noisy representation.

\subsection{Neural Networks}
Neural networks are a type of learning algorithm which forms the basis of most deep learning methods. A neural network comprises of neurons or units with some activation $a$ and parameters $\Theta = \{\mathcal{W}, \mathcal{B}\}$, where $\mathcal{W}$ is a set of weights and $\mathcal{B}$ a set of biases. The activation represents a linear combination of the input ${\bf x}$ to the neuron and the parameters, followed by an element-wise non-linearity $\sigma(\cdot)$, referred to as a transfer function: 
\begin{equation}
 a = \sigma({\bf w}^{T}{\bf x} + b).
\end{equation}
Typical transfer functions for traditional neural networks are the sigmoid and hyperbolic tangent function.  The multi-layered perceptrons (MLP), the most well-known of the traditional neural networks, have several layers of these transformations:
 \begin{equation}
 	f({\bf x}; \Theta) = \sigma( {\bf W}^{T}\sigma({\bf W}^{T} \ldots \sigma({\bf W}^{T} {\bf x} + b)  ) + b).
 \end{equation}
Here, ${\bf W}$ is a matrix comprising of columns ${\bf w}_{k}$, associated with activation $k$ in the output. Layers in between the input and output are often referred to as 'hidden' layers. When a neural network contains multiple hidden layers it is typically considered a 'deep' neural network, hence the term 'deep learning'.

At the final layer of the network the activations are mapped to a distribution over classes $P(y | {\bf x}; \Theta)$ through a {\it softmax} function:
\begin{equation}
 P(y | {\bf x}; \Theta) = \text{softmax}({\bf x}; \Theta) = \frac{e^{{\bf w}_{i}^{T}{\bf x} + b_{i}}}{\sum^{K}_{k = 1} e^{{\bf w}_{k}^{T}{\bf x} + b_{k}}},
\end{equation}
where ${\bf w}_{i}$ indicates the weight vector leading to the output node associated with class $i$. A schematic representation of three-layer MLP is shown in Figure \ref{fig:architectures}.

Maximum likelihood with stochastic gradient descent is currently the most popular method to fit parameters $\Theta$ to a dataset $\mathcal{D}$. In stochastic gradient descent a small subset of the data, a mini-batch, is used for each gradient update instead of the full data set. Optimizing maximum likelihood in practice amounts to minimizing the negative log-likelihood:
\begin{equation}
 \arg \min_{\Theta} - \sum^{N}_{n = 1} \log\big[ P(y_{n}| {\bf x}_{n}; \Theta) \big].
\end{equation}
This results in the binary cross-entropy loss for two-class problems and the categorical cross-entropy for multi-class tasks. A downside of this approach is that it typically does not optimize the quantity we are interested in directly, such as area under the receiver-operating characteristic (ROC) curve or common evaluation measures for segmentation, such as the Dice coefficient. 

For a long time, deep neural networks (DNN) were considered hard to train efficiently. They only gained popularity in 2006 \citep{Beng07,Hint06,Hint06a} when it was shown that training DNNs layer-by-layer in an unsupervised manner (pre-training), followed by supervised fine-tuning of the stacked network, could result in good performance. Two popular architectures trained in such a way are stacked auto-encoders (SAEs) and deep belief networks (DBNs). However, these techniques are rather complex and require a significant amount of engineering to generate satisfactory results.

Currently, the most popular models are trained end-to-end in a supervised fashion, greatly simplifying the training process. The most popular architectures are convolutional neural networks (CNNs) and recurrent neural networks (RNNs). CNNs are currently most widely used in (medical) image analysis, although RNNs are gaining popularity. The following sections will give a brief overview of each of these methods, starting with the most popular ones, and discussing their differences and potential challenges when applied to medical problems. 

\subsection{Convolutional Neural Networks (CNNs)}
\begin{figure*}[!tb]
\centering
\includegraphics[width=0.8\textwidth]{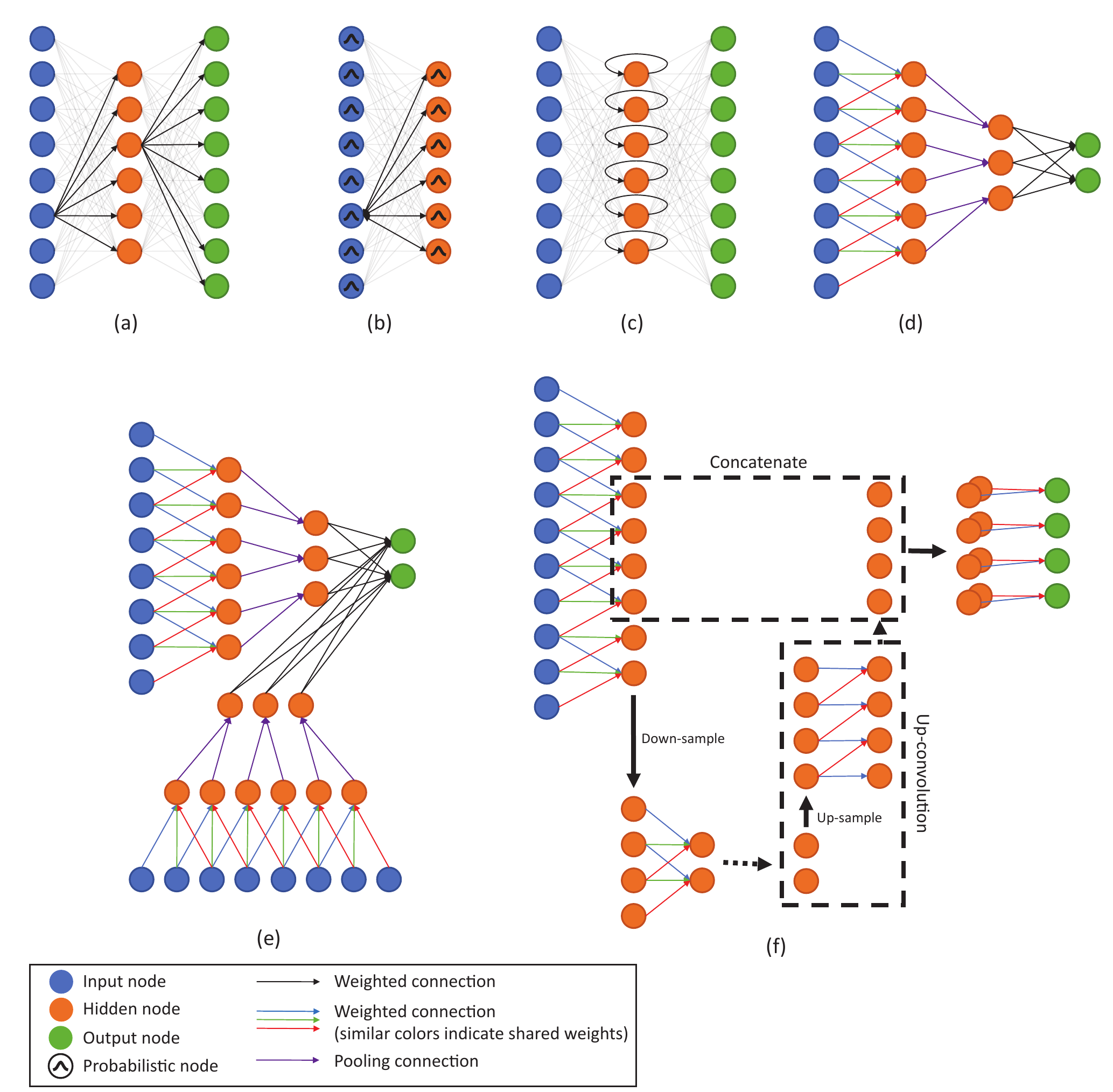}
\caption{Node graphs of 1D representations of architectures commonly used in medical imaging. a) Auto-encoder, b) restricted Boltzmann machine, c) recurrent neural network, d) convolutional neural network, e) multi-stream convolutional neural network, f) U-net (with a single downsampling stage). }.
\label{fig:architectures}
\end{figure*}

There are two key differences between MLPs and CNNs. First, in CNNs weights in the network are shared in such a way that it the network performs convolution operations on images. This way, the model does not need to learn separate detectors for the same object occurring at different positions in an image, making the network equivariant with respect to translations of the input. It also drastically reduces the amount of parameters (i.e.\ the number of weights no longer depends on the size of the input image) that need to be learned. An example of a 1D CNN is shown in Figure \ref{fig:architectures}.

At each layer, the input image is convolved with a set of $K$ kernels  $\mathcal{W} = \{ {\bf W}_{1}, {\bf W}_{2}, \ldots, {\bf W}_{K} \} $ and added biases $\mathcal{B} = \{b_{1}, \ldots, b_{K}\}$, each generating a new feature map ${\bf X}_{k}$. These features are subjected to an element-wise non-linear transform $\sigma(\cdot)$ and the same process is repeated for every convolutional layer $l$:
\begin{equation}
\label{eq::mapping_cnn}
 {\bf X}_{k}^{l} = \sigma\big( {\bf W}_{k}^{l -1} \ast {\bf X}^{l -1} + b_{k}^{l-1} \big).
\end{equation}

The second key difference between CNNs and MLPs, is the typical incorporation of pooling layers in CNNs, where pixel values of neighborhoods are aggregated using a permutation invariant function, typically the max or mean operation. This induces a certain amount of translation invariance and again reduces the amount of parameters in the network. At the end of the convolutional stream of the network, fully-connected layers (i.e.\ regular neural network layers) are usually added, where weights are no longer shared. Similar to MLPs, a distribution over classes is generated by feeding the activations in the final layer through a softmax function and the network is trained using maximum likelihood. 

\subsection{Deep CNN Architectures}
Given the prevalence of CNNs in medical image analysis, we elaborate on the most common architectures and architectural differences among the widely used models. 

\subsubsection{General classification architectures}
LeNet \citep{Lecu98} and AlexNet \citep{Kriz12}, introduced over a decade later, were in essence very similar models. Both networks were relatively shallow, consisting of two and five convolutional layers, respectively, and employed kernels with large receptive fields in layers close to the input and smaller kernels closer to the output. AlexNet did incorporate rectified linear units instead of the hyperbolic tangent as activation function.

After 2012 the exploration of novel architectures took off, and in the last three years there is a preference for far deeper models. By stacking smaller kernels, instead of using a single layer of kernels with a large receptive field, a similar function can be represented with less parameters. These deeper architectures generally have a lower memory footprint during inference, which enable their deployment on mobile computing devices such as smartphones. \cite{Simo14} were the first to explore much deeper networks, and employed small, fixed size kernels in each layer. A 19-layer model often referred to as VGG19 or OxfordNet won the ImageNet challenge of 2014. 

On top of the deeper networks, more complex building blocks have been introduced that improve the efficiency of the training procedure and again reduce the amount of parameters. \cite{Szeg14} introduced a 22-layer network named {\it GoogLeNet}, also referred to as Inception, which made use of so-called inception blocks \citep{Lin13c}, a module that replaces the mapping defined in Eq.~\eqref{eq::mapping_cnn} with a set of convolutions of different sizes. Similar to the stacking of small kernels, this allows a similar function to be represented with less parameters. The {\it ResNet} architecture \citep{He15b} won the ImageNet challenge in 2015 and consisted of so-called ResNet-blocks. Rather than learning a function, the residual block only learns the residual and is thereby pre-conditioned towards learning mappings in each layer that are close to the identity function. This way, even deeper models can be trained effectively. 

Since 2014, the performance on the ImageNet benchmark has saturated and it is difficult to assess whether the small increases in performance can really be attributed to 'better' and more sophisticated architectures. The advantage of the lower memory footprint these models provide is typically not as important for medical applications. Consequently, AlexNet or other simple models such as VGG are still popular for medical data, though recent landmark studies all use a version of GoogleNet called Inception v3 \citep{Guls16,Este17,Liu17}. Whether this is due to a superior architecture or simply because the model is a default choice in popular software packages is again difficult to assess. 

\subsubsection{Multi-stream architectures}
\label{sec:ms_architectures}
The default CNN architecture can easily accommodate multiple sources of information or representations of the input, in the form of channels presented to the input layer. This idea can be taken further and channels can be merged at any point in the network. Under the intuition that different tasks require different ways of fusion, {\it multi-stream} architectures are being explored. These models, also referred to as dual pathway architectures \citep{Kamn16}, have two main applications at the time of writing: (1) multi-scale image analysis and (2) 2.5D classification; both relevant for medical image processing tasks. 

For the detection of abnormalities, context is often an important cue. The most straightforward way to increase context is to feed larger patches to the network, but this can significantly increase the amount of parameters and memory requirements of a network. Consequently, architectures have been investigated where context is added in a down-scaled representation in addition to high resolution local information. To the best of our knowledge, the multi-stream multi-scale architecture was first explored by \cite{Fara13a}, who used it for segmentation in natural images. Several medical applications have also successfully used this concept \citep{Kamn16,Moes16,Song15,Yang16}. 

As so much methodology is still developed on natural images, the challenge of applying deep learning techniques to the medical domain often lies in adapting existing architectures to, for instance, different input formats such as three-dimensional data. In early applications of CNNs to such volumetric data, full 3D convolutions and the resulting large amount of parameters were circumvented by dividing the Volume of Interest (VOI) into slices which are fed as different streams to a network. \cite{Pras13} were the first to use this approach for knee cartilage segmentation. Similarly, the network can be fed with multiple angled patches from the 3D-space in a multi-stream fashion, which has been applied by various authors in the context of medical imaging \citep{Roth16,Seti16}. These approaches are also referred to as 2.5D classification. 

\subsubsection{Segmentation Architectures}
\label{sec:segm_arch}
Segmentation is a common task in both natural and medical image analysis and to tackle this, CNNs can simply be used to classify each pixel in the image individually, by presenting it with patches extracted around the particular pixel. A drawback of this naive 'sliding-window' approach is that input patches from neighboring pixels have huge overlap and the same convolutions are computed many times. Fortunately, the convolution and dot product are both linear operators and thus inner products can be written as convolutions and vice versa. By rewriting the fully connected layers as convolutions, the CNN can take input images larger than it was trained on and produce a likelihood map, rather than an output for a single pixel. The resulting 'fully convolutional network' (fCNN) can then be applied to an entire input image or volume in an efficient fashion. 

However, because of pooling layers, this may result in output with a far lower resolution than the input. 'Shift-and-stitch' \citep{Long15} is one of several methods proposed to prevent this decrease in resolution. The fCNN is applied to shifted versions of the input image. By stitching the result together, one obtains a full resolution version of the final output, minus the pixels lost due to the 'valid' convolutions. 

\cite{Ronn15} took the idea of the fCNN one step further and proposed the U-net architecture, comprising a 'regular' fCNN followed by an upsampling part where 'up'-convolutions are used to increase the image size, coined contractive and expansive paths. Although this is not the first paper to introduce learned upsampling paths in convolutional neural networks (e.g.~\cite{Long15}), the authors combined it with so called skip-connections to directly connect opposing contracting and expanding convolutional layers. A similar approach was used by \cite{Cice16} for 3D data. \cite{Mill16} proposed an extension to the U-Net layout that incorporates ResNet-like residual blocks and a Dice loss layer, rather than the conventional cross-entropy, that directly minimizes this commonly used segmentation error measure.

\subsection{Recurrent Neural Networks (RNNs)}
\label{sec:rnns}
Traditionally, RNNs were developed for discrete sequence analysis. They can be seen as a generalization of MLPs because both the input and output can be of varying length, making them suitable for tasks such as machine translation where a sentence of the source and target language are the input and output. In a classification setting, the model learns a distribution over classes $P(y | {\bf x}_{1}, {\bf x}_{2}, \ldots, {\bf x}_{T}; \Theta)$ given a sequence ${\bf x}_{1}, {\bf x}_{2}, \ldots, {\bf x}_{T}$, rather than a single input vector ${\bf x}$.

The plain RNN maintains a latent or hidden state ${\bf h}$ at time $t$ that is the output of a non-linear mapping from its input ${\bf x}_{t}$ and the previous state ${\bf h}_{t-1}$:
\begin{equation}
 {\bf h}_{t} = \sigma({\bf W}{\bf x}_{t} + {\bf R}{\bf h}_{t - 1} + {\bf b}),
\end{equation}
where weight matrices ${\bf W}$ and ${\bf R}$ are shared over time. For classification, one or more fully connected layers are typically added followed by a softmax to map the sequence to a posterior over the classes. 
\begin{equation}
 P(y | {\bf x}_{1}, {\bf x}_{2}, \ldots, {\bf x}_{T}; \Theta) = \text{softmax}( {\bf h}_{T}; {\bf W}_{out}, {\bf b}_{out}).
\end{equation}
Since the gradient needs to be backpropagated from the output through time, RNNs are inherently deep (in time) and consequently suffer from the same problems with training as regular deep neural networks \citep{Beng94}. To this end, several specialized memory units have been developed, the earliest and most popular being the Long Short Term Memory (LSTM) cell \citep{Hoch97}. The Gated Recurrent Unit \citep{Cho14} is a recent simplification of the LSTM and is also commonly used. \\

Although initially proposed for one-dimensional input, RNNs are increasingly applied to images. In natural images 'pixelRNNs' are used as autoregressive models, generative models that can eventually produce new images similar to samples in the training set. For medical applications, they have been used for segmentation problems, with promising results \citep{Stol15} in the MRBrainS challenge. 

\subsection{Unsupervised models}
\subsubsection{Auto-encoders (AEs) and Stacked Auto-encoders (SAEs)}
AEs are simple networks that are trained to reconstruct the input ${\bf x}$ on the output layer ${\bf x}'$ through one hidden layer ${\bf h}$. They are governed by a weight matrix ${\bf W}_{x, h}$ and bias $b_{x, h}$ from input to hidden state and ${\bf W}_{h, x'}$ with corresponding bias $b_{h, x'}$ from the hidden layer to the reconstruction. A non-linear function is used to compute the hidden activation:
\begin{equation}
\label{eq::ae_projection}
 {\bf h} = \sigma({\bf W}_{x, h} {\bf x} + {\bf b}_{x, h}).
\end{equation}
Additionally, the dimension of the hidden layer $|{\bf h}|$ is taken to be smaller than $|{\bf x}|$. This way, the data is projected onto a lower dimensional subspace representing a dominant latent structure in the input. Regularization or sparsity constraints can be employed to enhance the discovery process. If the hidden layer had the same size as the input and no further non-linearities were added, the model would simply learn the identity function.

The denoising auto-encoder \citep{Vinc10} is another solution to prevent the model from learning a trivial solution. Here the model is trained to reconstruct the input from a noise corrupted version (typically salt-and-pepper-noise). SAEs (or deep AEs) are formed by placing auto-encoder layers on top of each other. In medical applications surveyed in this work, auto-encoder layer were often trained individually (`greedily') after which the full network was fine-tuned using supervised training to make a prediction.

\subsubsection{Restricted Boltzmann Machines (RBMs) and Deep Belief Networks (DBNs)}
RBMs \citep{Hint10} are a type of Markov Random Field (MRF), constituting an input layer or visible layer ${\bf x} = (x_{1}, x_{2}, \ldots, x_{N})$ and a hidden layer ${\bf h} = (h_{1}, h_{2}, \ldots, h_{M})$ that carries the latent feature representation. The connections between the nodes are bi-directional, so given an input vector $\bf x$ one can obtain the latent feature representation $\bf h$ and also vice versa. As such, the RBM is a generative model, and we can sample from it and generate new data points. In analogy to physical systems, an energy function is defined for a particular state $({\bf x}, {\bf h})$ of input and hidden units:
\begin{equation}
 E({\bf x}, {\bf h}) = {\bf h}^{T}{\bf W}{\bf x} - {\bf c}^{T}{\bf x} - {\bf b}^{T}{\bf h},
\end{equation}
with ${\bf c}$ and ${\bf b}$ bias terms. The probability of the `state' of the system is defined by passing the energy to an exponential and normalizing:
\begin{equation}
 p({\bf x}, {\bf h}) = \frac{1}{Z} \exp\{ - E({\bf x}, {\bf h}) \}.
\end{equation}
Computing the partition function $Z$ is generally intractable. However, conditional inference in the form of computing ${\bf h}$ conditioned on ${\bf v}$ or vice versa is tractable and results in a simple formula:
\begin{equation}
 P(h_{j} | {\bf x}) = \frac{1}{ 1 + \exp\{ -b_{j} - {\bf W}_{j}{\bf x}\} }.
\end{equation}
Since the network is symmetric, a similar expression holds for $P(x_{i} | {\bf h})$.\\

DBNs \citep{Beng07,Hint06a} are essentially SAEs where the AE layers are replaced by RBMs. Training of the individual layers is, again, done in an unsupervised manner. Final fine-tuning is performed by adding a linear classifier to the top layer of the DBN and performing a supervised optimization. 

\subsubsection{Variational Auto-Encoders and Generative Adverserial Networks}
Recently, two novel unsupervised architectures were introduced: the variational auto-encoder (VAE) \citep{King13} and the generative adversarial network (GAN) \citep{Good14}. There are no peer-reviewed papers applying these methods to medical images yet, but applications in natural images are promising. We will elaborate on their potential in the discussion. 
\color{black}

\subsection{Hardware and Software}
One of the main contributors to steep rise of deep learning has been the widespread availability of GPU and GPU-computing libraries (CUDA, OpenCL). GPUs are highly parallel computing engines, which have an order of magnitude more execution threads than central processing units (CPUs). With current hardware, deep learning on GPUs is typically 10 to 30 times faster than on CPUs. 

Next to hardware, the other driving force behind the popularity of deep learning methods is the wide availability of open source software packages. These libraries provide efficient GPU implementations of important operations in neural networks, such as convolutions; allowing the user to implement ideas at a high level rather than worrying about low-level efficient implementations. At the time of writing, the most popular packages were (in alphabetical order):
\begin{itemize}
 \item {\bf Caffe} \citep{Jia14a}. Provides C++ and Python interfaces, developed by graduate students at UC Berkeley. 
 \item {\bf Tensorflow} \citep{Abad16}. Provides C++ and Python and interfaces, developed by Google and is used by Google research.
 \item {\bf Theano} \citep{Bast12}. Provides a Python interface, developed by MILA lab in Montreal.
 \item {\bf Torch} \citep{Coll11}. Provides a Lua interface and is used by, among others, Facebook AI research. 
\end{itemize}
There are third-party packages written on top of one or more of these frameworks, such as Lasagne (\url{https://github.com/Lasagne/Lasagne}) or Keras (\url{https://keras.io/}). It goes beyond the scope of this paper to discuss all these packages in detail.

\section{Deep Learning Uses in Medical Imaging}
\label{dl-mi}

\subsection{Classification}
\subsubsection{Image/exam classification}
Image or exam classification was one of the first areas in which deep learning made a major contribution to medical image analysis. In exam classification one typically has one or multiple images (an exam) as input with a single diagnostic variable as output (e.g., disease present or not). In such a setting, every diagnostic exam is a sample and dataset sizes are typically small compared to those in computer vision (e.g., hundreds/thousands vs.\ millions of samples). The popularity of transfer learning for such applications is therefore not surprising. 

Transfer learning is essentially the use of pre-trained networks (typically on natural images) to try to work around the (perceived) requirement of large data sets for deep network training. Two transfer learning strategies were identified: (1) using a pre-trained network as a feature extractor and (2) fine-tuning a pre-trained network on medical data. The former strategy has the extra benefit of not requiring one to train a deep network at all, allowing the extracted features to be easily plugged in to existing image analysis pipelines. Both strategies are popular and have been widely applied. However, few authors perform a thorough investigation in which strategy gives the best result. The two papers that do, \cite{Anto16} and \cite{Kim16a}, offer conflicting results. In the case of \cite{Anto16}, fine-tuning clearly outperformed feature extraction, achieving 57.6\% accuracy in multi-class grade assessment of knee osteoarthritis versus 53.4\%. \cite{Kim16a}, however, showed that using CNN as a feature extractor outperformed fine-tuning in cytopathology image classification accuracy (70.5\% versus 69.1\%). If any guidance can be given to which strategy might be most successful, we would refer the reader to two recent papers, published in high-ranking journals, which fine-tuned a pre-trained version of Google's Inception v3 architecture on medical data and achieved (near) human expert performance \citep{Este17,Guls16}. As far as the authors are aware, such results have not yet been achieved by simply using pre-trained networks as feature extractors.

With respect to the type of deep networks that are commonly used in exam classification, a timeline similar to computer vision is apparent. The medical imaging community initially focused on unsupervised pre-training and network architectures like SAEs and RBMs. The first papers applying these techniques for exam classification appeared in 2013 and focused on neuroimaging. \cite{Bros13}, \cite{Plis14}, \cite{Suk13}, and \cite{Suk14} applied DBNs and SAEs to classify patients as having Alzheimer's disease based on brain Magnetic Resonance Imaging (MRI). Recently, a clear shift towards CNNs can be observed. Out of the 47 papers published on exam classification in 2015, 2016, and 2017, 36 are using CNNs, 5 are based on AEs and 6 on RBMs. The application areas of these methods are very diverse, ranging from brain MRI to retinal imaging and digital pathology to lung computed tomography (CT). 

In the more recent papers using CNNs authors also often train their own network architectures from scratch instead of using pre-trained networks. \cite{Mene16} performed some experiments comparing training from scratch to fine-tuning of pre-trained networks and showed that fine-tuning worked better given a small data set of around a 1000 images of skin lesions. However, these experiments are too small scale to be able to draw any general conclusions from.

Three papers used an architecture leveraging the unique attributes of medical data: two use 3D convolutions \citep{Hoss16,Paya15} instead of 2D to classify patients as having Alzheimer; \cite{Kawa16b} applied a CNN-like architecture to a brain connectivity graph derived from MRI diffusion-tensor imaging (DTI). In order to do this, they developed several new layers which formed the basis of their network, so-called edge-to-edge, edge-to-node, and node-to-graph layers. They used their network to predict brain development and showed that they outperformed existing methods in assessing cognitive and motor scores.

Summarizing, in exam classification CNNs are the current standard techniques. Especially CNNs pre-trained on natural images have shown surprisingly strong results, challenging the accuracy of human experts in some tasks. Last, authors have shown that CNNs can be adapted to leverage intrinsic structure of medical images.

\subsubsection{Object or lesion classification}
\label{sec:object_classification}
Object classification usually focuses on the classification of a small (previously identified) part of the medical image into two or more classes (e.g. nodule classification in chest CT). For many of these tasks both local information on lesion appearance and global contextual information on lesion location are required for accurate classification. This combination is typically not possible in generic deep learning architectures. Several authors have used multi-stream architectures to resolve this in a multi-scale fashion (Section \ref{sec:ms_architectures}). \cite{Shen15b} used three CNNs, each of which takes a nodule patch at a different scale as input. The resulting feature outputs of the three CNNs are then concatenated to form the final feature vector. A somewhat similar approach was followed by \cite{Kawa16a} who used a multi-stream CNN to classify skin lesions, where each stream works on a different resolution of the image. \cite{Gao15} proposed to use a combination of CNNs and RNNs for grading nuclear cataracts in slit-lamp images, where CNN filters were pre-trained. This combination allows the processing of all contextual information regardless of image size.  
Incorporating 3D information is also often a necessity for good performance in object classification tasks in medical imaging. As images in computer vision tend to be 2D natural images, networks developed in those scenarios do not directly leverage 3D information. Authors have used different approaches to integrate 3D in an effective manner with custom architectures. \cite{Seti16} used a multi-stream CNN to classify points of interest in chest CT as a nodule or non-nodule. Up to nine differently oriented patches extracted from the candidate were used in separate streams and merged in the fully-connected layers to obtain the final classification output. In contrast, \cite{Nie16a} exploited the 3D nature of MRI by training a 3D CNN to assess survival in patients suffering from high-grade gliomas.

Almost all recent papers prefer the use of end-to-end trained CNNs. In some cases other architectures and approaches are used, such as RBMs \citep{Tuld16,Zhan16}, SAEs \citep{Chen16} and convolutional sparse auto-encoders (CSAE) \citep{Kall16}. The major difference between CSAE and a classic CNN is the usage of unsupervised pre-training with sparse auto-encoders. 

An interesting approach, especially in cases where object annotation to generate training data is expensive, is the integration of multiple instance learning (MIL) and deep learning. \cite{Xu14a} investigated the use of a MIL-framework with both supervised and unsupervised feature learning approaches as well as handcrafted features. The results demonstrated that the performance of the MIL-framework was superior to handcrafted features, which in turn closely approaches the performance of a fully supervised method. We expect such approaches to be popular in the future as well, as obtaining high-quality annotated medical data is challenging.

Overall, object classification sees less use of pre-trained networks compared to exam classifications, mostly due to the need for incorporation of contextual or three-dimensional information. Several authors have found innovative solutions to add this information to deep networks with good results, and as such we expect deep learning to become even more prominent for this task in the near future.

\subsection{Detection}
\subsubsection{Organ, region and landmark localization}
Anatomical object localization (in space or time), such as organs or landmarks, has been an important pre-processing step in segmentation tasks or in the clinical workflow for therapy planning and intervention. Localization in medical imaging often requires parsing of 3D volumes. To solve 3D data parsing with deep learning algorithms, several approaches have been proposed that treat the 3D space as a composition of 2D orthogonal planes. \cite{Yang15b} identified landmarks on the distal femur surface by processing three independent sets of 2D MRI slices (one for each plane) with regular CNNs. The 3D position of the landmark was defined as the intersection of the three 2D slices with the highest classification output. \cite{Vos16a} went one step further and localized regions of interest (ROIs) around anatomical regions (heart, aortic arch, and descending aorta) by identifying a rectangular 3D bounding box after 2D parsing the 3D CT volume. Pre-trained CNN architectures, as well as RBM, have been used for the same purpose \citep{Cai16,Chen15b,Kuma16}, overcoming the lack of data to learn better feature representations. All these studies cast the localization task as a classification task and as such generic deep learning architectures and learning processes can be leveraged.

Other authors try to modify the network learning process to directly predict locations. For example, \cite{Paye16} proposed to directly regress landmark locations with CNNs. They used landmark maps, where each landmark is represented by a Gaussian, as ground truth input data and the network is directly trained to predict this landmark map. Another interesting approach was published by \cite{Ghes16a}, in which reinforcement learning is applied to the identification of landmarks. The authors showed promising results in several tasks: 2D cardiac MRI and ultrasound (US) and 3D head/neck CT. 

Due to its increased complexity, only a few methods addressed the direct localization of landmarks and regions in the 3D image space. \cite{Zhen15a} reduced this complexity by decomposing 3D convolution as three one-dimensional convolutions for carotid artery bifurcation detection in CT data. \cite{Ghes16} proposed a sparse adaptive deep neural network powered by marginal space learning in order to deal with data complexity in the detection of the aortic valve in 3D transesophageal echocardiogram. 

CNNs have also been used for the localization of scan planes or key frames in temporal data. \cite{Baum16} trained CNNs on video frame data to detect up to 12 standardized scan planes in mid-pregnancy fetal US. Furthermore, they used saliency maps to obtain a rough localization of the object of interest in the scan plan (e.g. brain, spine). RNNs, particularly LSTM-RNNs, have also been used to exploit the temporal information contained in medical videos, another type of high dimensional data. \cite{Chen15d}, for example, employed LSTM models to incorporate temporal information of consecutive sequence in US videos for fetal standard plane detection. \cite{Kong16} combined an LSTM-RNN with a CNN to detect the end-diastole and end-systole frames in cine-MRI of the heart.

Concluding, localization through 2D image classification with CNNs seems to be the most popular strategy overall to identify organs, regions and landmarks, with good results. However, several recent papers expand on this concept by modifying the learning process such that accurate localization is directly emphasized, with promising results. We expect such strategies to be explored further as they show that deep learning techniques can be adapted to a wide range of localization tasks (e.g. multiple landmarks). RNNs have shown promise in localization in the temporal domain, and multi-dimensional RNNs could play a role in spatial localization as well.

\subsubsection{Object or lesion detection}
The detection of objects of interest or lesions in images is a key part of diagnosis and is one of the most labor-intensive for clinicians. Typically, the tasks consist of the localization and identification of small lesions in the full image space. There has been a long research tradition in computer-aided detection systems that are designed to automatically detect lesions, improving the detection accuracy or decreasing the reading time of human experts. Interestingly, the first object detection system using CNNs was already proposed in 1995, using a CNN with four layers to detect nodules in x-ray images \citep{Lo95}.

Most of the published deep learning object detection systems still uses CNNs to perform pixel (or voxel) classification, after which some form of post processing is applied to obtain object candidates. As the classification task performed at each pixel is essentially object classification, CNN architecture and methodology are very similar to those in section \ref{sec:object_classification}. The incorporation of contextual or 3D information is also handled using multi-stream CNNs (Section \ref{sec:ms_architectures}, for example by \cite{Barb16} and \cite{Roth16}. \cite{Tera16} used a multi-stream CNN to integrate CT and Positron Emission Tomography (PET) data. \cite{Dou16} used a 3D CNN to find micro-bleeds in brain MRI. Last, as the annotation burden to generate training data can be similarly significant compared to object classification, weakly-supervised deep learning has been explored by \cite{Hwan16a}, who adopted such a strategy for the detection of nodules in chest radiographs and lesions in mammography.

There are some aspects which are significantly different between object detection and object classification. One key point is that because every pixel is classified, typically the class balance is skewed severely towards the non-object class in a training setting. To add insult to injury, usually the majority of the non-object samples are easy to discriminate, preventing the deep learning method to focus on the challenging samples. \cite{Grin16b} proposed a selective data sampling in which wrongly classified samples were fed back to the network more often to focus on challenging areas in retinal images. Last, as classifying each pixel in a sliding window fashion results in orders of magnitude of redundant calculation, fCNNs, as used in \cite{Wolt16}, are important aspect of an object detection pipeline as well.

Challenges in meaningful application of deep learning algorithms in object detection are thus mostly similar to those in object classification. Only few papers directly address issues specific to object detection like class imbalance/hard-negative mining or efficient pixel/voxel-wise processing of images. We expect that more emphasis will be given to those areas in the near future, for example in the application of multi-stream networks in a fully convolutional fashion.

\subsection{Segmentation}
\subsubsection{Organ and substructure segmentation}
The segmentation of organs and other substructures in medical images allows quantitative analysis of clinical parameters related to volume and shape, as, for example, in cardiac or brain analysis. Furthermore, it is often an important first step in computer-aided detection pipelines. The task of segmentation is typically defined as identifying the set of voxels which make up either the contour or the interior of the object(s) of interest. Segmentation is the most common subject of papers applying deep learning to medical imaging (Figure \ref{fig:statistics}), and as such has also seen the widest variety in methodology, including the development of unique CNN-based segmentation architectures and the wider application of RNNs. 

The most well-known, in medical image analysis, of these novel CNN architectures is U-net, published by \cite{Ronn15} (section \ref{sec:segm_arch}). The two main architectural novelties in U-net are the combination of an equal amount of upsampling and downsampling layers. Although learned upsampling layers have been proposed before, U-net combines them with so-called skip connections between opposing convolution and deconvolution layers. This which concatenate features from the contracting and expanding paths. From a training perspective this means that entire images/scans can be processed by U-net in one forward pass, resulting in a segmentation map directly. This allows U-net to take into account the full context of the image, which can be an advantage in contrast to patch-based CNNs. Furthermore, in an extended paper by \cite{Cice16}, it is shown that a full 3D segmentation can be achieved by feeding U-net with a few 2D annotated slices from the same volume. Other authors have also built derivatives of the U-net architecture; \cite{Mill16}, for example, proposed a 3D-variant of U-net architecture, called V-net, performing 3D image segmentation using 3D convolutional layers with an objective function directly based on the Dice coefficient. \cite{Droz16} investigated the use of short ResNet-like skip connections in addition to the long skip-connections in a regular U-net.

RNNs have recently become more popular for segmentation tasks. For example, \cite{Xie16a} used a spatial clockwork RNN to segment the perimysium in H\&E-histopathology images. This network takes into account prior information from both the row and column predecessors of the current patch. To incorporate bidirectional information from both left/top and right/bottom neighbors, the RNN is applied four times in different orientations and the end-result is concatenated and fed to a fully-connected layer. This produces the final output for a single patch. \cite{Stol15} where the first to use a 3D LSTM-RNN with convolutional layers in six directions. \cite{Ande16} used a 3D RNN with gated recurrent units to segment gray and white matter in a brain MRI data set. \cite{Chen16h} combined bi-directional LSTM-RNNs with 2D U-net-like-architectures to segment structures in anisotropic 3D electron microscopy images. Last, \cite{Poud16} combined a 2D U-net architecture with a gated recurrent unit to perform 3D segmentation.

Although these specific segmentation architectures offered compelling advantages, many authors have also obtained excellent segmentation results with patch-trained neural networks. One of the earliest papers covering medical image segmentation with deep learning algorithms used such a strategy and was published by \cite{Cire12b}. They applied pixel-wise segmentation of membranes in electron microscopy imagery in a sliding window fashion. Most recent papers now use fCNNs (subsection \ref{sec:segm_arch}) in preference over sliding-window-based classification to reduce redundant computation.

fCNNs have also been extended to 3D and have been applied to multiple targets at once: \cite{Kore16}, used 3D fCNNs to generate vertebral body likelihood maps which drove deformable models for vertebral body segmentation in MR images, \cite{Zhou16} segmented nineteen targets in the human torso, and \cite{Moes16a} trained a single fCNN to segment brain MRI, the pectoral muscle in breast MRI, and the coronary arteries in cardiac CT angiography (CTA). 

One challenge with voxel classification approaches is that they sometimes lead to spurious responses. To combat this, groups have tried to combine fCNNs with graphical models like MRFs \citep{Shak16, Song15} and Conditional Random Fields (CRFs) \citep{Alan16,Cai16a,Chri16,Dou16,Fu16a,Gao16b} to refine the segmentation output. In most of the cases, graphical models are applied on top of the likelihood map produced by CNNs or fCNNs and act as label regularizers.

Summarizing, segmentation in medical imaging has seen a huge influx of deep learning related methods. Custom architectures have been created to directly target the segmentation task. These have obtained promising results, rivaling and often improving over results obtained with fCNNs. 

\subsubsection{Lesion segmentation}
Segmentation of lesions combines the challenges of object detection and organ and substructure segmentation in the application of deep learning algorithms. Global and local context are typically needed to perform accurate segmentation, such that multi-stream networks with different scales or non-uniformly sampled patches are used as in for example \cite{Kamn16} and \cite{Ghaf16}. In lesion segmentation we have also seen the application of U-net and similar architectures to leverage both this global and local context. The architecture used by \cite{Wang15d}, similar to the U-net, consists of the same downsampling and upsampling paths, but does not use skip connections. Another U-net-like architecture was used by \cite{Bros16} to segment white matter lesions in brain MRI. However, they used 3D convolutions and a single skip connection between the first convolutional and last deconvolutional layers.

One other challenge that lesion segmentation shares with object detection is class imbalance, as most voxels/pixels in an image are from the non-diseased class. Some papers combat this by adapting the loss function: \cite{Bros16} defined it to be a weighted combination of the sensitivity and the specificity, with a larger weight for the specificity to make it less sensitive to the data imbalance. Others balance the data set by performing data augmentation on positive samples \citep{Kamn16,Litj16c,Pere16}.

Thus lesion segmentation sees a mixture of approaches used in object detection and organ segmentation. Developments in these two areas will most likely naturally propagate to lesion segmentation as the existing challenges are also mostly similar.

\subsection{Registration}
Registration (i.e.\ spatial alignment) of medical images is a common image analysis task in which a coordinate transform is calculated from one medical image to another. Often this is performed in an iterative framework where a specific type of (non-)parametric transformation is assumed and a pre-determined metric (e.g.\ L2-norm) is optimized. Although segmentation and lesion detection are more popular topics for deep learning, researchers have found that deep networks can be beneficial in getting the best possible registration performance. Broadly speaking, two strategies are prevalent in current literature: (1) using deep-learning networks to estimate a similarity measure for two images to drive an iterative optimization strategy, and (2) to directly predict transformation parameters using deep regression networks.

\cite{Wu13c}, \cite{Simo16}, and \cite{Chen15f} used the first strategy to try to optimize registration algorithms. \cite{Chen15f} used two types of stacked auto-encoders to assess the local similarity between CT and MRI images of the head. Both auto-encoders take vectorized image patches of CT and MRI and reconstruct them through four layers. After the networks are pre-trained using unsupervised patch reconstruction they are fine-tuned using two prediction layers stacked on top of the third layer of the SAE. These prediction layers determine whether two patches are similar (class 1) or dissimilar (class 2). \cite{Simo16} used a similar strategy, albeit with CNNs, to estimate a similarity cost between two patches from differing modalities. However, they also presented a way to use the derivative of this metric to directly optimize the transformation parameters, which are decoupled from the network itself. Last, \cite{Wu13c} combined independent subspace analysis and convolutional layers to extract features from input patches in an unsupervised manner. The resultant feature vectors are used to drive the HAMMER registration algorithm instead of handcrafted features. 

\cite{Miao16a} and \cite{Yang16a} used deep learning algorithms to directly predict the registration transform parameters given input images. \cite{Miao16a} leveraged CNNs to perform 3D model to 2D x-ray registration to assess the pose and location of an implanted object during surgery. In total the transformation has 6 parameters, two translational, 1 scaling and 3 angular parameters. They parameterize the feature space in steps of 20 degrees for two angular parameters and train a separate CNN to predict the update to the transformation parameters given an digitally reconstructed x-ray of the 3D model and the actual inter-operative x-ray. The CNNs are trained with artificial examples generated by manually adapting the transformation parameters for the input training data. They showed that their approach has significantly higher registration success rates than using traditional - purely intensity based - registration methods. \cite{Yang16a} tackled the problem of prior/current registration in brain MRI using the OASIS data set. They used the large deformation diffeomorphic metric mapping (LDDMM) registration methodology as a basis. This method takes as input an initial momentum value for each pixel which is then evolved over time to obtain the final transformation. However, the calculation of the initial momentum map is often an expensive procure. The authors circumvent this by training a U-net like architecture to predict the x- and y-momentum map given the input images. They obtain visually similar results but with significantly improved execution time: 1500x speed-up for 2D and 66x speed-up for 3D.

In contrast to classification and segmentation, the research community seems not have yet settled on the best way to integrate deep learning techniques in registration methods. Not many papers have yet appeared on the subject and existing ones each have a distinctly different approach. Thus, giving recommendations on what method is most promising seems inappropriate. However, we expect to see many more contributions of deep learning to medical image registration in the near future.

\subsection{Other tasks in medical imaging}
\subsubsection{Content-based image retrieval}
Content-based image retrieval (CBIR) is a technique for knowledge discovery in massive databases and offers the possibility to identify similar case histories, understand rare disorders, and, ultimately, improve patient care. The major challenge in the development of CBIR methods is extracting effective feature representations from the pixel-level information and associating them with meaningful concepts. The ability of deep CNN models to learn rich features at multiple levels of abstraction has elicited interest from the CBIR community.

All current approaches use (pre-trained) CNNs to extract feature descriptors from medical images. \cite{Anav16} and \cite{Liu16a} applied their methods to databases of X-ray images. Both used a five-layer CNN and extracted features from the fully-connected layers. \cite{Anav16} used the last layer and a pre-trained network. Their best results were obtained by feeding these features to a one-vs-all support vector machine (SVM) classifier to obtain the distance metric. They showed that incorporating gender information resulted in better performance than just CNN features. \cite{Liu16a} used the penultimate fully-connected layer and a custom CNN trained to classify X-rays in 193 classes to obtain the descriptive feature vector. After descriptor binarization and data retrieval using Hamming separation values, the performance was inferior to the state of the art, which the authors attributed to small patch sizes of 96 pixels. The method proposed by \cite{Shah16} combines CNN feature descriptors with hashing-forests. 1000 features were extracted for overlapping patches in prostate MRI volumes, after which a large feature matrix was constructed over all volumes. Hashing forests were then used to compress this into descriptors for each volume.

Content-based image retrieval as a whole has thus not seen many successful applications of deep learning methods yet, but given the results in other areas it seems only a matter of time. An interesting avenue of research could be the direct training of deep networks for the retrieval task itself.

\begin{figure}[!tb]
\centering
\includegraphics[width=0.48\textwidth]{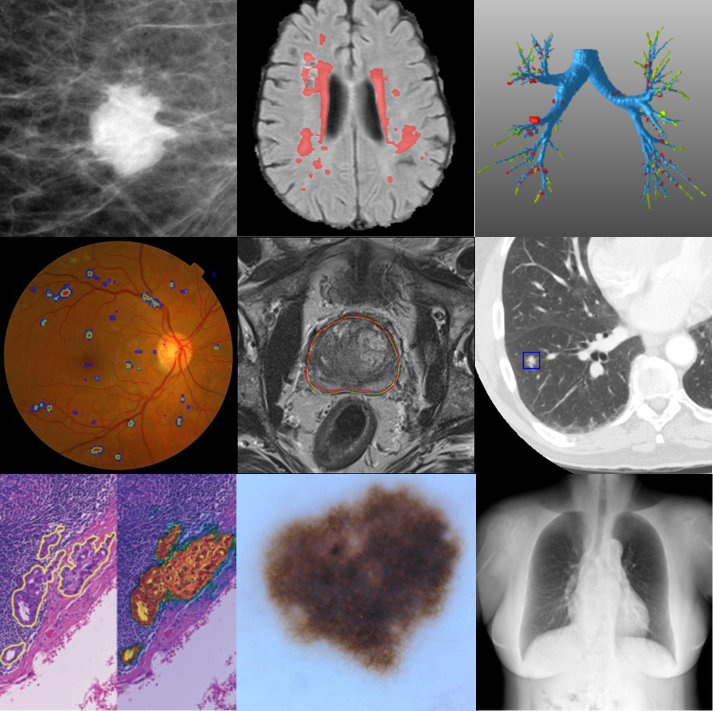}
\caption{Collage of some medical imaging applications in which deep learning has achieved state-of-the-art results. From top-left to bottom-right: mammographic mass classification \citep{Kooi16}, segmentation of lesions in the brain (top ranking in BRATS, ISLES and MRBrains challenges, image from \cite{Ghaf16}, leak detection in airway tree segmentation \citep{Char16c}, diabetic retinopathy classification (Kaggle Diabetic Retinopathy challenge 2015, image from \cite{Grin16b}, prostate segmentation (top rank in PROMISE12 challenge), nodule classification (top ranking in LUNA16 challenge), breast cancer metastases detection in lymph nodes (top ranking and human expert performance in CAMELYON16), human expert performance in skin lesion classification \citep{Este17}, and state-of-the-art bone suppression in x-rays, image from \cite{Yang16}.
\label{fig:dl_applications}}
\end{figure}

\begin{table*}[!tb]
  \centering
  \caption{Overview of papers using deep learning techniques for brain image analysis. All works use MRI unless otherwise mentioned.}
    \resizebox{\textwidth}{!}{\begin{tabular}{llllp{6cm}}
    \toprule
    Reference & Method & Application; remarks \\
    \midrule
    \multicolumn{3}{l}{Disorder classification (AD, MCI, Schizophrenia)}\\
    \midrule
\cite{Bros13} 	& DBN 	&  	AD/HC classification; Deep belief networks with convolutional RBMs for manifold learning\\
\cite{Plis14} 	& DBN 	& 	Deep belief networks evaluated on brain network estimation, Schizophrenia and Huntington's disease classification\\
\cite{Suk13} 	& SAE 	&  	AD/MCI classification; Stacked auto encoders with supervised fine tuning\\
\cite{Suk14} 	& RBM 	&  	AD/MCI/HC classification; Deep Boltzmann Machines on MRI and PET modalities\\
\cite{Paya15} 	& CNN 	&  	AD/MCI/HC classification; 3D CNN pre-trained with sparse auto-encoders\\
\cite{Suk15} 	& SAE 	&  	AD/MCI/HC classification; SAE for latent feature extraction on a large set of hand-crafted features from MRI and PET\\
\cite{Hoss16} 	& CNN 	&  	AD/MCI/HC classification; 3D CNN pre-trained with a 3D convolutional auto-encoder on fMRI data\\
\cite{Kim16b} 	& ANN 	&  	Schizophrenia/NH classification on fMRI; Neural network showing advantage of pre-training with SAEs, and L1 sparsification\\
\cite{Orti16} 	& DBN 	&  	AD/MCI/HC classification; An ensemble of Deep belief networks, with their votes fused using an SVM classifier\\
\cite{Pina16} 	& DBN 	&  	Schizophrenia/NH classification; DBN pre-training followed by supervised fine-tuning\\
\cite{Sarr16} 	& CNN 	&  	AD/HC classification; Adapted Lenet-5 architecture on fMRI data\\
\cite{Suk16} 	& SAE 	& 	MCI/HC classification of fMRI data; Stacked auto-encoders for feature extraction, HMM as a generative model on top\\
\cite{Suk16a} 	& CNN 	&  	AD/MCI/HC classification; CNN on sparse representations created by regression models\\
\cite{Shi17} 	& ANN 	&  	AD/MCI/HC classification; Multi-modal stacked deep polynomial networks with an SVM classifier on top using MRI and PET\\

    \midrule
    \multicolumn{3}{l}{Tissue/anatomy/lesion/tumor segmentation}\\
    \midrule
\cite{Guo14a} 	& SAE 	& 	Hippocampus segmentation; SAE for representation learning used for target/atlas patch similarity measurement\\
\cite{Breb15} 	& CNN 	& 	Anatomical segmentation; fusing multi-scale 2D patches with a 3D patch using a CNN\\
\cite{Choi16a} 	& CNN 	& 	Striatum segmentation; Two-stage (global/local) approximations with 3D CNNs\\
\cite{Stol15} 	& RNN 	& 	Tissue segmentation; PyraMiD-LSTM, best brain segmentation results on MRBrainS13 (and competitive results on EM-ISBI12)\\
\cite{Zhan15} 	& CNN 	& 	Tissue segmentation; multi-modal 2D CNN\\
\cite{Ande16} 	& RNN 	& 	Tissue segmentation; two convolutional gated recurrent units in different directions for each dimension\\
\cite{Bao16} 	& CNN 	& 	Anatomical segmentation; Multi-scale late fusion CNN with random walker as a novel label consistency method\\
\cite{Bire16} 	& CNN 	& 	Lesion segmentation; Multi-view (2.5D) CNN concatenating features from previous time step for a longitudinal analysis\\
\cite{Bros16} 	& CNN 	& 	Lesion segmentation; Convolutional encoder-decoder network with shortcut connections and convolutional RBM pretraining \\
\cite{Chen16g} 	& CNN 	& 	Tissue segmentation; 3D res-net combining features from different layers\\
\cite{Ghaf16} 	& CNN 	& 	Lesion segmentation; CNN trained on non-uniformly sampled patch to integrate a larger context with a foviation effect\\
\cite{Ghaf16a} 	& CNN 	& 	Lesion segmentation; multi-scale CNN with late fusion that integrates anatomical location information into network\\
\cite{Hava16a} 	& CNN 	& 	Tumor segmentation; CNN handling missing modalities with abstraction layer that transforms feature maps to their statistics\\
\cite{Hava16} 	& CNN 	& 	Tumor segmentation; two-path way CNN with different receptive fields\\
\cite{Kamn16} 	& CNN 	& 	Tumor segmentation; 3D multi-scale fully convolutional network with CRF for label consistency\\
\cite{Klee16} 	& CNN 	& 	Brain extraction; 3D fully convolutional CNN on multi-modal input\\  
\cite{Mans16} 	& SAE 	& 	Visual pathway segmentation; Learning appearance features from SAE for steering the shape model for segmentation\\
\cite{Mill16a} 	& CNN 	& 	Anatomical segmentation on MRI and US; Hough-voting to acquire mapping from CNN features to full patch segmentations\\
\cite{Moes16} 	& CNN 	& 	Tissue segmentation; CNN trained on multiple patch sizes\\
\cite{Nie16} 	& CNN 	& 	Infant tissue segmentation; FCN with a late fusion method on different modalities\\
\cite{Pere16} 	& CNN 	& 	Tumor segmentation; CNN on multiple modality input\\
\cite{Shak16} 	& CNN 	& 	Anatomical segmentation; FCN followed by Markov random fields\\
\cite{Zhao16} 	& CNN 	& 	Tumor segmentation; Multi-scale CNN with a late fusion architecture\\
    \midrule
    \multicolumn{3}{l}{Lesion/tumor detection and classification}\\
    \midrule

\cite{Pan15} 	& CNN 	& 	Tumor grading; 2D tumor patch classification using a CNN\\
\cite{Dou15} 	& ISA 	& 	Microbleed detection; 3D stacked Independent Subspace Analysis for candidate feature extraction, SVM classification \\  
\cite{Dou16} 	& CNN 	& 	Microbleed detection; 3D FCN for candidate segmentation followed by a 3D CNN as false positive reduction \\
\cite{Ghaf17} 	& CNN 	& 	Lacune detection; FCN for candidate segmentation then a multi-scale 3D CNN with anatomical features as false positive reduction\\

    \midrule
    \multicolumn{3}{l}{Survival/disease activity/development prediction}\\
    \midrule

\cite{Kawa16b} 	& CNN 	& 	Neurodevelopment prediction; CNN with specially-designed edge-to-edge, edge-to-node and node-to-graph conv. layers for brain nets\\
\cite{Nie16a} 	& CNN 	& 	Survival prediction; features from a Multi-modal 3D CNN is fused with hand-crafted features to train an SVM\\
\cite{Yoo16} 	& CNN 	& 	Disease activity prediction; Training a CNN on the Euclidean distance transform of the lesion masks as the input\\
\cite{Burg17} 	& CNN 	& 	Survival prediction; DBN on MRI and fusing it with clinical characteristics and structural connectivity data\\
    \midrule
    \multicolumn{3}{l}{Image construction/enhancement}\\
    \midrule

\cite{Li14a} 	& CNN 	& 	Image construction; 3D CNN for constructing PET from MR images\\
\cite{Bahr16} 	& CNN 	& 	Image construction; 3D CNN for constructing 7T-like images from 3T MRI\\
\cite{Beno16} 	& SAE 	& 	Denoising DCE-MRI; using an ensemble of denoising SAE (pretrained with RBMs)\\
\cite{Golk16} 	& CNN 	&   Image construction; Per-pixel neural network to predict complex diffusion parameters based on fewer measurements\\
\cite{Hoff16a} 	& ANN 	& 	Image construction; Deep neural nets with SRelu nonlinearity for thermal image construction\\
\cite{Nie16b} 	& CNN 	& 	Image construction; 3D fully convolutional network for constructing CT from MR images\\
\cite{Seve16} 	& ANN 	& 	Image construction; Encoder-decoder network for synthesizing one MR modality from another\\

    \midrule
    \multicolumn{3}{l}{Other}\\
    \midrule
\cite{Bros14} 	& DBN 	& 	Manifold Learning; DBN with conv. RBM layers for modeling the variability in brain morphology and lesion distribution in MS\\
\cite{Chen15f} 	& ANN 	& 	Similarity measurement; neural network fusing the moving and reference image patches, pretrained with SAE\\
\cite{Huan16} 	& RBM 	& 	fMRI blind source separation; RBM for both internal and functional interaction-induced latent sources detection\\
\cite{Simo16} 	& CNN 	& 	Similarity measurement; 3D CNN estimating similarity between reference and moving images stacked in the input\\
\cite{Wu13c} 	& ISA 	& 	Correspondence detection in deformable registration; stacked convolutional ISA for unsupervised feature learning \\
\cite{Yang16a} 	& CNN 	& 	Image registration; Conv. encoder-decoder net. predicting momentum in x and y directions, given the moving and fixed image patches\\

\bottomrule    
    \end{tabular}}%
  \label{tab:brain}%
\end{table*}%

\subsubsection{Image Generation and Enhancement}
A variety of image generation and enhancement methods using deep architectures have been proposed, ranging from removing obstructing elements in images, normalizing images, improving image quality, data completion, and pattern discovery.

In image generation, 2D or 3D CNNs are used to convert one input image into another. Typically these architectures lack the pooling layers present in classification networks. These systems are then trained with a data set in which both the input and the desired output are present, defining the differences between the generated and desired output as the loss function. Examples are regular and bone-suppressed X-ray in \cite{Yang16}, 3T and 7T brain MRI in \cite{Bahr16}, PET from MRI in \cite{Li14a}, and CT from MRI in \cite{Nie16b}. \cite{Li14a} even showed that one can use these generated images in computer-aided diagnosis systems for Alzheimer's disease when the original data is missing or not acquired. 

With multi-stream CNNs super-resolution images can be generated from multiple low-resolution inputs (section \ref{sec:ms_architectures}). In \cite{Okta16}, multi-stream networks reconstructed high-resolution cardiac MRI from one or more low-resolution input MRI volumes. Not only can this strategy be used to infer missing spatial information, but can also be leveraged in other domains; for example, inferring advanced MRI diffusion parameters from limited data \citep{Golk16}.  Other image enhancement applications like intensity normalization and denoising have seen only limited application of deep learning algorithms. \cite{Jano16} used SAEs to normalize H\&E-stained histopathology images whereas \cite{Beno16} used CNNs to perform denoising in DCE-MRI time-series.

Image generation has seen impressive results with very creative applications of deep networks in significantly differing tasks. One can only expect the number of tasks to increase further in the future.

\begin{table*}[htb!]
  \centering
  \caption{Overview of papers using deep learning techniques for retinal image analysis. All works use CNNs.}
    \resizebox{\textwidth}{!}{\begin{tabular}{llp{6cm}}
    \toprule
\multicolumn{2}{l}{Color fundus images: segmentation of anatomical structures and quality assessment}\\
\midrule
\cite{Fu16} & Blood vessel segmentation; CNN combined with CRF to model long-range pixel interactions\\  
\cite{Fu16a} & Blood vessel segmentation; extending the approach by \cite{Fu16} by reformulating CRF as RNN\\
\cite{Maha16} & Image quality assessment; classification output using CNN-based features combined with the output using saliency maps\\
\cite{Mani16} & Segmentation of blood vessels and optic disk; VGG-19 network extended with specialized layers for each segmentation task\\
\cite{Wu16} & Blood vessel segmentation; patch-based CNN followed by mapping PCA solution of last layer feature maps to full segmentation\\
\cite{Zill16} & Segmentation of the optic disk and the optic cup; simple CNN with filters sequentially learned using boosting\\

\midrule
\multicolumn{2}{l}{Color fundus images: detection of abnormalities and diseases}\\
\midrule
\cite{Chen15a} & Glaucoma detection; end-to-end CNN, the input is a patch centered at the optic disk\\
\cite{Abra16} & Diabetic retinopathy detection; end-to-end CNN, outperforms traditional method, evaluated on a public dataset\\
\cite{Burl16} & Age-related macular degeneration detection; uses overfeat pretrained network for feature extraction\\
\cite{Grin16b} & Hemorrhage detection; CNN dynamically trained using selective data sampling to perform hard negative mining\\
\cite{Guls16} & Diabetic retinopathy detection; Inception network, performance comparable to a panel of  seven certified ophthalmologists\\
\cite{Pren16a} & Hard exudate detection; end-to-end CNN combined with the outputs of traditional classifiers for detection of landmarks\\  
\cite{Worr16} & Retinopathy of prematurity detection; fine-tuned ImageNet trained GoogLeNet, feature map visualization to highlight disease\\

\midrule
\multicolumn{2}{l}{Work in other imaging modalities}\\
\midrule
\cite{Gao15} & Cataract classification in slit lamp images; CNN followed by a set of recursive neural networks to extract higher order features\\
\cite{Schl15} & Fluid segmentation in OCT; weakly supervised CNN improved with semantic descriptors from clinical reports\\
\cite{Pren16} & Blood vessel segmentation in OCT angiography; simple CNN, segmentation of several capillary networks\\
    \bottomrule    
    \end{tabular}}%
  \label{tab:eye}%
\end{table*}%

\subsubsection{Combining Image Data With Reports}
The combination of text reports and medical image data has led to two avenues of research: (1) leveraging reports to improve image classification accuracy \citep{Schl15}, and (2) generating text reports from images \citep{Kisi16,Shin15,Shin16a,Wang16b}; the latter inspired by recent caption generation papers from natural images \citep{Karp15}. To the best of our knowledge, the first step towards leveraging reports was taken by \cite{Schl15}, who argued that large amounts of annotated data may be difficult to acquire and proposed to add semantic descriptions from reports as labels. The system was trained on sets of images along with their textual descriptions and was taught to predict semantic class labels during test time. They showed that semantic information increases classification accuracy for a variety of pathologies in Optical Coherence Tomography (OCT) images.

\cite{Shin15} and \cite{Wang16b} mined semantic interactions between radiology reports and images from a large data set extracted from a PACS system. They employed latent Dirichlet allocation (LDA), a type of stochastic model that generates a distribution over a vocabulary of topics based on words in a document. In a later work, \cite{Shin16a} proposed a system to generate descriptions from chest X-rays. A CNN was employed to generate a representation of an image one label at a time, which was then used to train an RNN to generate sequence of MeSH keywords. \cite{Kisi16} used a completely different approach and predicted categorical BI-RADS descriptors for breast lesions. In their work they focused on three descriptors used in mammography: shape, margin, and density, where each have their own class label. The system was fed with the image data and region proposals and predicts the correct label for each descriptor (e.g. for shape either oval, round, or irregular). 

Given the wealth of data that is available in PACS systems in terms of images and corresponding diagnostic reports, it seems like an ideal avenue for future deep learning research. One could expect that advances in captioning natural images will in time be applied to these data sets as well.

\section{Anatomical application areas}
\label{sec:app}
This section presents an overview of deep learning contributions to the various application areas in medical imaging. We highlight some key contributions and discuss performance of systems on large data sets and on public challenge data sets. All these challenges are listed on \url{http:\\\\www.grand-challenge.org}. 

\subsection{Brain}
DNNs have been extensively used for brain image analysis in several different application domains (Table \ref{tab:brain}). A large number of studies address classification of Alzheimer's disease and segmentation of brain tissue and anatomical structures (e.g. the hippocampus). Other important areas are detection and segmentation of lesions (e.g. tumors, white matter lesions, lacunes, micro-bleeds).

Apart from the methods that aim for a scan-level classification (e.g.\ Alzheimer diagnosis), most methods learn mappings from local patches to representations and subsequently from representations to labels. However, the local patches might lack the contextual information required for tasks where anatomical information is paramount (e.g. white matter lesion segmentation). To tackle this, \cite{Ghaf16} used non-uniformly sampled patches by gradually lowering sampling rate in patch sides to span a larger context. An alternative strategy used by many groups is multi-scale analysis and a fusion of representations in a fully-connected layer. 

Even though brain images are 3D volumes in all surveyed studies, most methods work in 2D, analyzing the 3D volumes slice-by-slice. This is often motivated by either the reduced computational requirements or the thick slices relative to in-plane resolution in some data sets. More recent publications had also employed 3D networks.

DNNs have completely taken over many brain image analysis challenges. In the 2014 and 2015 brain tumor segmentation challenges (BRATS), the 2015 longitudinal multiple sclerosis lesion segmentation challenge, the 2015 ischemic stroke lesion segmentation challenge (ISLES), and the 2013 MR brain image segmentation challenge (MRBrains), the top ranking teams to date have all used CNNs. Almost all of the aforementioned methods are concentrating on brain MR images. We expect that other brain imaging modalities such as CT and US can also benefit from deep learning based analysis.

\begin{table*}[htb]
  \centering
  \caption{Overview of papers using deep learning techniques for chest x-ray image analysis.}
    \resizebox{\textwidth}{!}{\begin{tabular}{llllp{6cm}}
    \toprule
    Reference & Application & Remarks \\
    \midrule
\cite{Lo95}		& Nodule detection 					& Classifies candidates from small patches with two-layer CNN, each with 12 $5\times5$ filters \\
\cite{Anav15}	& Image retrieval					& Combines classical features with those from pre-trained CNN for image retrieval using SVM \\
\cite{Bar15}	& Pathology detection				& Features from a pre-trained CNN and low level features are used to detect various diseases \\
\cite{Anav16}	& Image retrieval					& Continuation of \cite{Anav15}, adding age and gender as features \\
\cite{Bar16}	& Pathology detection				& Continuation of \cite{Bar15}, more experiments and adding feature selection \\
\cite{Cice16a}	& Pathology detection				& GoogLeNet CNN detects five common abnormalities, trained and validated on a large data set \\ 
\cite{Hwan16}	& Tuberculosis detection			& Processes entire radiographs with a pre-trained fine-tuned network with 6 convolution layers \\
\cite{Kim16}	& Tuberculosis detection			& MIL framework produces heat map of suspicious regions via deconvolution \\
\cite{Shin16a}	& Pathology detection				& CNN detects 17 diseases, large data set (7k images), recurrent networks produce short captions \\
\cite{Rajk16}	& Frontal/lateral classification	& Pre-trained CNN performs frontal/lateral classification task \\
\cite{Yang16}	& Bone suppression					& Cascade of CNNs at increasing resolution learns bone images from gradients of radiographs \\
\cite{Wang16e}	& Nodule classification				& Combines classical features with CNN features from pre-trained ImageNet CNN \\
\bottomrule    
\end{tabular}}%
\label{tab:cxr}%
\end{table*}%

\begin{table*}[htb!]
  \centering
  \caption{Overview of papers using deep learning techniques for chest CT image analysis.}
    \resizebox{\textwidth}{!}{\begin{tabular}{llllp{6cm}}
    \toprule
    Reference & Application; remarks \\
    \midrule

\multicolumn{2}{l}{Segmentation}\\
\midrule
\cite{Char16c}&Airway segmentation where multi-view CNN classifies candidate branches as true airways or leaks\\
\midrule
\multicolumn{2}{l}{Nodule detection and analysis}\\
\midrule
\cite{Ciom15a}	& Used a standard feature extractor and a pre-trained CNN to classify detected lesions as benign peri-fissural nodules \\ 
\cite{Ginn15}	& Detects nodules with pre-trained CNN features from orthogonal patches around candidate, classified with SVM \\
\cite{Shen15b}	& Three CNNs at different scales estimate nodule malignancy scores of radiologists (LIDC-IDRI data set) \\
\cite{Chen16b}	& Combines features from CNN, SDAE and classical features to characterize nodules from LIDC-IDRI data set \\
\cite{Ciom16}	& Multi-stream CNN to classify nodules into subtypes: solid, part-solid, non-solid, calcified, spiculated, perifissural \\
\cite{Dou16b}	& Uses 3D CNN around nodule candidates; ranks \#1 in LUNA16 nodule detection challenge \\ 
\cite{Li16a}	& Detects nodules with 2D CNN that processes small patches around a nodule \\
\cite{Seti16}	& Detects nodules with end-to-end trained multi-stream CNN with 9 patches per candidate \\ 
\cite{Shen16}	& 3D CNN classifies volume centered on nodule as benign/malignant, results are combined to patient level prediction \\
\cite{Sun16a}	& Same dataset as \cite{Shen15b}, compares CNN, DBN, SDAE and classical computer-aided diagnosis schemes \\
\cite{Tera16}	& Combines features extracted from 2 orthogonal CT patches and a PET patch \\ 
\midrule
\multicolumn{2}{l}{Interstitial lung disease}\\
\midrule
\cite{Anth16}	& Classification of 2D patches into interstitial lung texture classes using a standard CNN\\
\cite{Chri16a}	& 2D interstitial pattern classification with CNNs pre-trained with a variety of texture data sets\\
\cite{Gao16b}	& Propagates manually drawn segmentations using CNN and CRF for more accurate interstitial lung disease reference\\ 
\cite{Gao16d}	& AlexNet applied to large parts of 2D CT slices to detect presence of interstitial patterns\\ 
\cite{Gao16c}	& Uses regression to predict area covered in 2D slice with a particular interstitial pattern\\
\cite{Tara16}	& Combines existing computer-aided diagnosis system and CNN to classify lung texture patterns.\\
\cite{Tuld16}	& Classification of lung texture and airways using an optimal set of filters derived from DBNs and RBMs\\
\midrule
\multicolumn{2}{l}{Other applications}\\
\midrule
\cite{Tajb15a} 	& Multi-stream CNN to detect pulmonary embolism from candidates obtained from a tobogganing algorithm\\
\cite{Carn16}	& Predicts 5-year mortality from thick slice CT scans and segmentation masks\\
\cite{Vos16}	& Identifies the slice of interest and determine the distance between CT slices\\
   \bottomrule    
    \end{tabular}}%
  \label{tab:chestct}%
\end{table*}%

\subsection{Eye}
\label{sec:eye}
Ophthalmic imaging has developed rapidly over the past years, but only recently are deep learning algorithms being applied to eye image understanding. As summarized in Table \ref{tab:eye}, most works employ simple CNNs for the analysis of color fundus imaging (CFI). A wide variety of applications are addressed: segmentation of anatomical structures, segmentation and detection of retinal abnormalities, diagnosis of eye diseases, and image quality assessment. 

In 2015, Kaggle organized a diabetic retinopathy detection competition: Over 35,000 color fundus images were provided to train algorithms to predict the severity of disease in 53,000 test images. The majority of the 661 teams that entered the competition applied deep learning and four teams achieved performance above that of humans, all using end-to-end CNNs. Recently \cite{Guls16} performed a thorough analysis of the performance of a Google Inception v3 network for diabetic retinopathy detection, showing performance comparable to a panel of seven certified ophthalmologists.

\renewcommand*\footnoterule{}
\setlength{\skip\footins}{2cm}

\begin{table*}[htb!]
\begin{minipage}{\textwidth}
  \centering
  \caption{Overview of papers using deep learning for digital pathology images. The staining and imaging modality abbreviations used in the table are as follows: H\&E: hematoxylin and eosin staining, TIL: Tumor-infiltrating lymphocytes, BCC: Basal cell carcinoma, IHC: immunohistochemistry, RM: Romanowsky, EM: Electron microscopy, PC: Phase contrast, FL: Fluorescent, IFL: Immunofluorescent, TPM: Two-photon microscopy, CM: Confocal microscopy, Pap: Papanicolaou.}
    \resizebox{\textwidth}{!}{\begin{tabular}{lllllp{6cm}}
    \toprule
    Reference & Topic & Staining\textbackslash Modality & Method \\
\midrule
\multicolumn{2}{l}{Nucleus detection, segmentation, and classification}\\
\midrule
\cite{Cire13a} 	& Mitosis detection 					& H\&E 			& CNN-based pixel classifier\\
\cite{Cruz13a} 	& Detection of basal cell carcinoma						& H\&E 			& Convolutional auto-encoder neural network\\
\cite{Malo13} 	& Mitosis detection 					& H\&E 			& Combines shape‑based features with CNN\\
\cite{Wang14c} 	& Mitosis detection 					& H\&E 			& Cascaded ensemble of CNN and handcrafted features\\
\cite{Ferr15} 	& Bacterial colony counting 			& Culture plate & CNN-based patch classifier\\
\cite{Ronn15} 	& Cell segmentation 					& EM 			& U-Net with deformation augmentation\\
\cite{Shko15} 	& Mitosis detection 					& Live-imaging 	& CNN-based patch classifier\\
\cite{Song15} 	& Segmentation of cytoplasm and nuclei 	& H\&E 			& Multi-scale CNN and graph-partitioning-based method\\
\cite{Xie15} 	& Nucleus detection 					& Ki-67 		& CNN model that learns the voting offset vectors and voting confidence\\
\cite{Xie15a} 	& Nucleus detection 					& H\&E, Ki-67 	& CNN-based structured regression model for cell detection\\
\cite{Akra16} 	& Cell segmentation 					& FL, PC, H\&E 	& fCNN for cell bounding box proposal and  CNN for segmentation\\
\cite{Alba16} 	& Mitosis detection 					& H\&E 			& Incorporated `crowd sourcing' layer into the CNN framework\\
\cite{Baue16} 	& Nucleus classification 				& IHC 			& CNN-based patch classifier\\
\cite{Chen16a} 	& Mitosis detection 					& H\&E 			& Deep regression network (DRN)\\
\cite{Gao16}	& Nucleus classification 				& IFL 			& Classification of Hep2-cells with CNN\\
\cite{Han16} 	& Nucleus classification 				& IFL 			& Classification of Hep2-cells with CNN\\
\cite{Jano16b} 	& Nucleus segmentation 					& H\&E 			& Resolution adaptive deep hierarchical learning scheme \\
\cite{Kash16} 	& Nucleus detection 					& H\&E 			& Combination of CNN and hand-crafted features\\
\cite{Mao16} 	& Mitosis detection 					& PC 			& Hierarchical CNNs for patch sequence classification\\
\cite{Mish16} 	& Classification of mitochondria 		& EM 			& CNN-based patch classifier\\
\cite{Phan16} 	& Nucleus classification 				& FL 			& Classification of Hep2-cells using transfer learning (pre-trained CNN)\\
\cite{Romo16} 	& Tubule nuclei detection				& H\&E 			& CNN-based classification of pre-selected candidate nuclei\\
\cite{Siri16} 	& Nucleus detection and classification 	& H\&E 			& CNN with spatially constrained regression\\
\cite{Song16} 	& Cell segmentation  					& H\&E 			& Multi-scale CNN\\
\cite{Turk16} 	& TIL detection 						& H\&E 			& CNN-based classification of superpixels\\
\cite{Veta16} 	& Nuclear area measurement 				& H\&E 			& A CNN directly measures nucleus area without requiring segmentation \\
\cite{Wang16c} 	& Subtype cell detection				& H\&E 			& Combination of two CNNs for joint cell detection and classification\\
\cite{Xie16} 	& Nucleus detection and cell counting 	& FL and H\&E 	& Microscopy cell counting with fully convolutional regression networks\\
\cite{Xing16} 	& Nucleus segmentation 					& H\&E, IHC 	& CNN and selection-based sparse shape model \\
\cite{Xu16b} 	& Nucleus detection 					& H\&E 			& Stacked sparse auto-encoders (SSAE)\\
\cite{Xu16e} 	& Nucleus detection 					& Various		& General deep learning framework to detect cells in whole-slide images\\
\cite{Yang16c} 	& Glial cell segmentation 				& TPM 			& fCNN with an iterative k-terminal
cut algorithm\\
\cite{Yao16} 	& Nucleus classification 				& H\&E 			& Classifies cellular tissue into tumor, lymphocyte, and stromal\\
\cite{Zhao16b} 	& Classification of leukocytes 			& RM 		 	& CNN-based patch classifier\\

\midrule
\multicolumn{2}{l}{Large organ segmentation}\\ 
\midrule
\cite{Cire12b} 	& Segmentation of neuronal membranes 		& EM 		& Ensemble of several CNNs with different architectures\\
\cite{Kain15} 	& Segmentation of colon glands 				& H\&E 		& Used two CNNs to segment glands and their separating structures \\
\cite{Apou16} 	& Detection of lobular structures in breast & IHC 		& Combined the outputs of a CNN and a texture classification system\\
\cite{BenT16a} 	& Segmentation of colon glands 				& H\&E 		& fCNN with a loss accounting for smoothness and object interactions\\
\cite{BenT16} 	& Segmentation of colon glands 				& H\&E 		& A multi-loss fCNN to perform both segmentation and classification\\
\cite{Chen16h} 	& Neuronal membrane and fungus segmentation & EM 		& Combination of bi-directional LSTM-RNNs and kU-Nets\\
\cite{Chen16d} 	& Segmentation of colon glands 				& H\&E 		& Deep contour-aware CNN\\
\cite{Cice16} 	& Segmentation of xenopus kidney 			& CM 		& 3D U-Net \\
\cite{Droz16} 	& Segmentation of neuronal structures 		& EM 		& fCNN with skip connections\\
\cite{Li16} 	& Segmentation of colon glands 				& H\&E 		& Compares CNN with an SVM using hand-crafted features\\
\cite{Teik16} 	& Volumetric vascular segmentation 			& FL  		& Hybrid 2D-3D CNN architecture\\
\cite{Wang16} 	& Segmentation of messy and muscle regions 	& H\&E 		& Conditional random field jointly trained with an fCNN\\
\cite{Xie16a} 	& Perimysium segmentation 					& H\&E 		& 2D spatial clockwork RNN\\
\cite{Xu16a} 	& Segmentation of colon glands 				& H\&E 		& Used three CNNs to predict gland and contour pixels\\
\cite{Xu16d} 	& Segmenting epithelium \& stroma 			& H\&E, IHC & CNNs applied to over-segmented image regions (superpixels)\\

\midrule
\multicolumn{2}{l}{Detection and classification of disease}\\     
\midrule
\cite{Cruz14} 	& Detection of invasive ductal carcinoma 		& H\&E 				& CNN-based patch classifier\\
\cite{Xu14a} 	& Patch-level classification of colon cancer 	& H\&E 				& Multiple instance learning framework with CNN features\\
\cite{Bych16} 	& Outcome prediction of colorectal cancer		& H\&E 				& Extracted CNN features from epithelial tissue for prediction\\
\cite{Chan17} 	& Multiple cancer tissue classification 		& Various 			& Transfer learning using multi-Scale convolutional sparse coding\\
\cite{Gunh16} 	& Grading glioma 								& H\&E 				& Ensemble of CNNs\\
\cite{Kaell16} 	& Predicting Gleason score 						& H\&E 				& OverFeat pre-trained network as feature extractor\\
\cite{Kim16a} 	& Thyroid cytopathology classification 			& H\&E, RM \& Pap 	& Fine-tuning pre-trained AlexNet\\
\cite{Litj16c} 	& Detection of prostate and breast cancer		& H\&E 			   	& fCNN-based pixel classifier\\
\cite{Quin16} 	& Malaria, tuberculosis and parasites detection & Light microscopy 	& CNN-based patch classifier\\
\cite{Reza16} 	& Gleason grading and breast cancer detection 	& H\&E 				& The system incorporates shearlet features inside a CNN\\
\cite{Scha16a} 	& SPOP mutation prediction of prostate cancer 	& H\&E 				& Ensemble of ResNets\\
\cite{Wang16a} 	& Metastases detection in lymph node 			& H\&E 				& Ensemble of CNNs with hard negative mining\\
 
\midrule
\multicolumn{2}{l}{Other pathology applications}\\     
\midrule
\cite{Jano16} 	& Stain normalization 						& H\&E 		& Used SAE for classifying tissue and subsequent histogram matching\\
\cite{Jano16a} 	& Deep learning tutorial 					& Various 	& Covers different detecting, segmentation, and classification tasks\\
\cite{Seth16} 	& Comparison of normalization algorithms 	& H\&E 		& Presents effectiveness of stain normalization for application of CNNs \\
\bottomrule    
    \end{tabular}}%
  \label{tab:pathology}%
  \end{minipage}

\end{table*}%

\subsection{Chest}
In thoracic image analysis of both radiography and computed tomography, the detection, characterization, and classification of nodules is the most commonly addressed application. Many works add features derived from deep networks to existing feature sets or compare CNNs with classical machine learning approaches using handcrafted features. In chest X-ray, several groups detect multiple diseases with a single system. In CT the detection of textural patterns indicative of interstitial lung diseases is also a popular research topic. 

Chest radiography is the most common radiological exam; several works use a large set of images with text reports to train systems that combine CNNs for image analysis and RNNs for text analysis. This is a branch of research we expect to see more of in the near future.

In a recent challenge for nodule detection in CT, LUNA16, CNN architectures were used by all top performing systems. This is in contrast with a previous lung nodule detection challenge, ANODE09, where handcrafted features were used to classify nodule candidates. The best systems in LUNA16 still rely on nodule candidates computed by rule-based image processing, but systems that use deep networks for candidate detection also performed very well (e.g.\ U-net). Estimating the probability that an individual has lung cancer from a CT scan is an important topic: It is the objective of the Kaggle Data Science Bowl 2017, with \$1 million in prizes and more than one thousand participating teams.

\begin{table*}[htb!]
  \centering
  \caption{Overview of papers using deep learning techniques for breast image analysis. MG = mammography; TS = tomosynthesis; US = ultrasound; ADN = Adaptive Deconvolution Network.}
    \resizebox{\textwidth}{!}{\begin{tabular}{lllllp{6cm}}
    \toprule
    Reference & Modality & Method & Application; remarks \\
    \midrule
    \cite{Sahi96} 	& MG				& CNN & First application of a CNN to mammography \\
    \cite{Jami12} 	& MG, US 			& ADN & Four layer ADN, an early form of CNN for mass classification\\     
    \cite{Fons15} 	& MG				& CNN &	Pre-trained network extracted features classified with SVM for breast density estimation \\
    \cite{Akse16} 	& MG				& CNN &	Use a modified region proposal CNN (R-CNN) for the localization and classification of masses \\
	\cite{Arev16}	& MG				& CNN & Lesion classification, combination with hand-crafted features gave the best performance \\
	\cite{Dalm17}   & MRI 				& CNN & Breast and fibroglandular tissue segmentation \\    
    \cite{Dubr16}   & MG				& CNN & Tissue classification using regular CNNs \\
    \cite{Dhun16} 	& MG				& CNN &	Combination of different CNNs combined with hand-crafted features \\
    \cite{Foti16} 	& TS				& CNN &	Improved state-of-the art for mass detection in tomosynthesis \\
    \cite{Hwan16a} 	& MG				& CNN & Weakly supervised CNN for localization of masses \\
    \cite{Huyn16}	& MG				& CNN & Pre-trained CNN on natural image patches applied to mass classification \\	
    \cite{Kall16} 	& MG				& SAE &	Unsupervised CNN feature learning with SAE for breast density classification \\
    \cite{Kisi16} 	& MG				& CNN & R-CNN combined with multi-class loss trained on semantic descriptions of potential masses \\
    \cite{Kooi16} 	& MG				& CNN &	Improved the state-of-the art for mass detection and show human performance on a patch level \\
    \cite{Qiu16a} 	& MG				& CNN &	CNN for direct classification of future risk of developing cancer based on negative mammograms \\
    \cite{Sama16} 	& TS				& CNN &	Microcalcification detection \\
    \cite{Sama16a} 	& TS				& CNN & Pre-trained CNN on mammographic masses transfered to tomosynthesis \\
    \cite{Sun16}  	& MG				& CNN & Semi-supervised CNN for classification of masses \\
    \cite{Zhan16} 	& US 				& RBM & Classification benign vs. malignant with shear wave elastography\\
    \cite{Kooi17}  	& MG				& CNN & Pre-trained CNN on mass/normal patches to discriminate malignant masses from (benign) cysts \\    
    \cite{Wang17} 	& MG 				& CNN & Detection of cardiovascular disease based on vessel calcification \\    
    \bottomrule    
    \end{tabular}}%
  \label{tab:breast}%
\end{table*}
\subsection{Digital pathology and microscopy}
The growing availability of large scale gigapixel whole-slide images (WSI) of tissue specimen has made digital pathology and microscopy a very popular application area for deep learning techniques. The developed techniques applied to this domain focus on three broad challenges: (1) Detecting, segmenting, or classifying nuclei, (2) segmentation of large organs, and (3) detecting and classifying the disease of interest at the lesion- or WSI-level. Table \ref{tab:pathology} presents an overview for each of these categories.

Deep learning techniques have also been applied for normalization of histopathology images. Color normalization is an important research area in histopathology image analysis. In \cite{Jano16}, a method for stain normalization of hematoxylin and eosin (H\&E) stained histopathology images was presented based on deep sparse auto-encoders. Recently, the importance of color normalization was demonstrated by \cite{Seth16} for CNN based tissue classification in H\&E stained images.

The introduction of grand challenges in digital pathology has fostered the development of computerized digital pathology techniques. The challenges that evaluated existing and new approaches for analysis of digital pathology images are: EM segmentation challenge 2012 for the 2D segmentation of neuronal processes, mitosis detection challenges in ICPR 2012 and AMIDA 2013, GLAS for gland segmentation and, CAMELYON16 and TUPAC for processing breast cancer tissue samples.

In both ICPR 2012 and the AMIDA13 challenges on mitosis detection the IDSIA team outperformed other algorithms with a CNN based approach \citep{Cire13a}. The  same team had the highest performing system in EM 2012 \citep{Cire12b} for 2D segmentation of neuronal processes. In their approach, the task of segmenting membranes of neurons was performed by mild smoothing and thresholding of the output of a CNN, which computes pixel probabilities. 

GLAS addressed the problem of gland instance segmentation in colorectal cancer tissue samples. \cite{Xu16a} achieved the highest rank using three CNN models. The first CNN classifies pixels as gland versus non-gland. From each feature map of the first CNN, edge information is extracted using the holistically nested edge technique, which uses side convolutions to produce an edge map. Finally, a third CNN merges gland and edge maps to produce the final segmentation.

CAMELYON16 was the first challenge to provide participants with WSIs. 
Contrary to other medical imaging applications, the availability of large amount of annotated data in this challenge allowed for training very deep models such as 22-layer GoogLeNet \citep{Szeg14}, 16-layer VGG-Net \citep{Simo14}, and 101-layer ResNet \citep{He15b}. The top-five performing systems used one of these architectures. The best performing solution in the Camelyon16 challenge was presented in \cite{Wang16a}. This method is based on an ensemble of two GoogLeNet architectures, one trained with and one without hard-negative mining to tackle the challenge. The latest submission of this team using the WSI standardization algorithm by \cite{Ehte16} achieved an AUC of 0.9935, for task 2, which outperformed the AUC of a pathologist (AUC = 0.966) who independently scored the complete test set.

The recently held TUPAC challenge addressed detection of mitosis in breast cancer tissue, and prediction of tumor grading at the WSI level. The top performing system by \cite{Paen16} achieved the highest performance in all tasks. The method has three main components: (1) Finding high cell density regions, (2) using a CNN to detect mitoses in the regions of interest, (3) converting the results of mitosis detection to a feature vector for each WSI and using an SVM classifier to compute the tumor proliferation and molecular data scores.
\begin{table*}[!tb]
  \centering
  \caption{Overview of papers using deep learning techniques for cardiac image analysis.}
    \resizebox{\textwidth}{!}{\begin{tabular}{lllllp{6cm}}
    \toprule
    Reference & Modality & Method & Application; remarks \\
    \midrule
\cite{Emad15} 	& MRI & CNN 		& Left ventricle slice detection; simple CNN indicates if structure is present\\  
\cite{Aven16} 	& MRI & CNN 		& Left ventricle segmentation; AE used to initialize filters because training data set was small\\
\cite{Kong16} 	& MRI & RNN 		& Identification of end-diastole and end-systole frames from cardiac sequences\\
\cite{Okta16} 	& MRI & CNN 		& Super-resolution; U-net/ResNet hybrid, compares favorably with standard superresolution methods\\
\cite{Poud16} 	& MRI & RNN 		& Left ventricle segmentation; RNN processes stack of slices, evaluated on several public datasets\\
\cite{Rupp16} 	& MRI & CNN 		& Cardiac structure segmentation; patch-based CNNs integrated in active contour framework\\
\cite{Tran16} 	& MRI & CNN 		& Left and right ventricle segmentation; 2D fCNN architecture, evaluated on several public data sets\\
\cite{Yang16b}	& MRI & CNN 		& Left ventricle segmentation; CNN combined with multi-atlas segmentation\\
\cite{Zhan16b} 	& MRI & CNN 		& Identifying presence of apex and base slices in cardiac exam for quality assessment\\
\cite{Ngo17} 	& MRI & DBN 		& Left ventricle segmentation; DBN is used to initialize a level set framework\\
\midrule
\cite{Carn12} 	& US & DBN 			& Left ventricle segmentation; DBN embedded in system using landmarks and non-rigid registration\\
\cite{Carn13} 	& US & DBN 			& Left ventricle tracking; extension of \cite{Carn12} for tracking\\
\cite{Chen16e} 	& US & CNN 			& Structure segmentation in 5 different 2D views; uses transfer learning\\
\cite{Ghes16} 	& US & CNN 			& 3D aortic valve detection and segmentation; uses shallow and deeper sparse networks\\
\cite{Nasc16} 	& US & DBN 			& Left ventricle segmentation; DBN applied to patches steers multi-atlas segmentation process\\
\cite{Mora16b} 	& US & CNN 			& Automatic generation of text descriptions for Doppler US images of cardiac valves using doc2vec\\
\midrule
\cite{Guel16} 	& CT & CNN 			& Coronary centerline extraction; CNN classifies paths as correct or leakages\\
\cite{Less16} 	& CT & CNN 			& Coronary calcium detection in low dose ungated CT using multi-stream CNN (3 views)\\
\cite{Mora16} 	& CT & CNN 			& Labeling of 2D slices from cardiac CT exams; comparison with handcrafted features\\
\cite{Vos16a} 	& CT & CNN 			& Detect bounding boxes by slice classification and combining 3 orthogonal 2D CNNs\\
\cite{Wolt16} 	& CT & CNN 			& Coronary calcium detection in gated CTA; compares 3D CNN with multi-stream 2D CNNs\\
\cite{Zrei16} 	& CT & CNN 			& Left ventricle segmentation; multi-stream CNN (3 views) voxel classification\\
    \bottomrule    
    \end{tabular}}%
  \label{tab:cardiac}%
\end{table*}%

\subsection{Breast}
One of the earliest DNN applications from \cite{Sahi96} was on breast imaging. Recently, interest has returned which resulted in significant advances over the state of the art, achieving the performance of human readers on ROIs \citep{Kooi16}. Since most breast imaging techniques are two dimensional, methods successful in natural images can easily be transferred. With one exception, the only task addressed is the detection of breast cancer; this consisted of three subtasks: (1) detection and classification of mass-like lesions, (2) detection and classification of micro-calcifications, and (3) breast cancer risk scoring of images.  Mammography is by far the most common modality and has consequently enjoyed the most attention. Work on tomosynthesis, US, and shear wave elastography is still scarce, and we have only one paper that analyzed breast MRI with deep learning; these other modalities will likely receive more attention in the next few years. Table \ref{tab:breast} summarizes the literature and main messages. 

Since many countries have screening initiatives for breast cancer, there should be massive amounts of data available, especially for mammography, and therefore enough opportunities for deep models to flourish. Unfortunately, large public digital databases are unavailable and consequently older scanned screen-film data sets are still in use. Challenges such as the recently launched DREAM challenge have not yet had the desired success.

As a result, many papers used small data sets resulting in mixed performance. Several projects have addressed this issue by exploring semi-supervised learning \citep{Sun16}, weakly supervised learning \citep{Hwan16a}, and transfer learning \citep{Kooi17,Sama16a}). Another method combines deep models with handcrafted features \citep{Dhun16}, which have been shown to be complementary still, even for very big data sets \citep{Kooi16}. State of the art techniques for mass-like lesion detection and classification tend to follow a two-stage pipeline with a candidate detector; this design reduces the image to a set of potentially malignant lesions, which are fed to a deep CNN \citep{Foti16, Kooi16}. Alternatives use a region proposal network (R-CNN) that bypasses the cascaded approach \citep{Akse16, Kisi16}. 

When large data sets are available, good results can be obtained. At the SPIE Medical Imaging conference of 2016, a researcher from a leading company in the mammography CAD field told a packed conference room how a few weeks of experiments with a standard architecture (AlexNet) - trained on the company's proprietary database - yielded a performance that was superior to what years of engineering handcrafted feature systems had achieved \citep{Foti16}. 

\begin{table*}[htb!]
  \centering
  \caption{Overview of papers using deep learning for abdominal image analysis.}
    \resizebox{\textwidth}{!}{\begin{tabular}{lllllp{6cm}}
    \toprule
    Reference & Topic & Modality & Method & Remarks \\
\midrule
{\em Multiple} &&&&\\
\midrule
\cite{Hu16b}	& Segmentation	& CT & CNN & 3D CNN with time-implicit level sets for segmentation of liver, spleen and kidneys \\
\midrule
\multicolumn{3}{l}{\em Segmentation tasks in liver imaging}\\    
\midrule
\cite{Li15b} 	& Lesion & CT & CNN & 2D 17$\times$17 patch-based classification, \cite{Ben16} repeats this approach \\
\cite{Viva15}   & Lesion & CT & CNN & 2D CNN for liver tumor segmentation in follow-up CT taking baseline CT as input \\
\cite{Ben16} 	& Liver & CT & CNN & 2D CNN similar to U-net, but without cross-connections; good results on SLIVER07\\
\cite{Chri16} 	& Liver \& tumor & CT & CNN & U-net, cascaded fCNN and dense 3D CRF\\
\cite{Dou16a} 	& Liver & CT & CNN & 3D CNN \(\)with conditional random field; good results on SLIVER07\\
\cite{Hoog16}	& Lesion & CT/MRI & CNN & 2D CNN obtained probabilities are used to drive active contour model\\
\cite{Hu16a}	& Liver & CT & CNN & 3D CNN with surface evolution of a shape prior; good results on SLIVER07 \\ 
\cite{Lu17} 	& Liver & CT & CNN & 3D CNN, competitive results on SLIVER07 \\
\midrule
{\em Kidneys} &&&&\\    
\midrule
\cite{Lu16a} 	& Localization & CT & CNN 	& Combines local patch and slice based CNN\\
\cite{Ravi16a} 	& Localization & US & CNN 	& Combines CNN with classical features to detect regions around kidneys\\
\cite{Thon16} 	& Segmentation & CT & CNN 	& 2D CCN with 43$\times$43 patches, tested on 20 scans\\
\midrule
\multicolumn{3}{l}{\em Pancreas segmentation in CT} \\    
\midrule
\cite{Fara15}	&Segmentation	&CT	&CNN & Approach with elements similar to \cite{Roth15d}\\ 
\cite{Roth15d}	&Segmentation	&CT	&CNN & Orthogonal patches from superpixel regions are fed into CNNs in three different ways\\ 
\cite{Cai16a} 	&Segmentation 	&CT &CNN & 2 CNNs detect inside and boundary of organ, initializes conditional random field\\ 
\cite{Roth16a}	&Segmentation 	&CT &CNN & 2 CNNs detect inside and boundary of pancreas, combined with random forests\\ 
\midrule
{\em Colon} &&&&\\    
\midrule
\cite{Tajb15}& Polyp detection & Colonoscopy & CNN& CNN computes additional features, improving existing scheme\\ 
\cite{Liu16}& Colitis detection & CT & CNN& Pre-trained ImageNet CNN generates features for linear SVM\\ 
\cite{Napp16}& Polyp detection & CT &CNN& Substantial reduction of false positives using pre-trained and fine-tuned CNN\\ 
\cite{Tach16}& Electronic cleansing &CT &CNN& Voxel classification in dual energy CT, material other than soft tissue is removed\\ 
\cite{Zhan17}& Polyp detection & Colonoscopy & CNN & Pre-trained ImageNet CNN for feature extraction, two SVMs for cascaded classification\\
\midrule
\multicolumn{3}{l}{\em Prostate segmentation in MRI}\\    
\midrule
\cite{Liao13} &\multicolumn{4}{l}{Application of stacked independent subspace analysis networks} \\    
\cite{Chen16c} &\multicolumn{4}{l}{CNN produces energy map for 2D slice based active appearance segmentation}\\ 
\cite{Guo16} & \multicolumn{4}{l}{Stacked sparse auto-encoders extract features from patches, input to atlas matching and a deformable model}\\ 
\cite{Mill16}&\multicolumn{4}{l}{3D U-net based CNN architecture with objective function that directly optimizes Dice coefficient, ranks \#5 in PROMISE12}\\ 
\cite{Lequ17} & \multicolumn{4}{l}{3D fully convolutional network, hybrid between a ResNet and U-net architecture, ranks \#1 on PROMISE12}\\
\midrule
{\em Prostate} &&&&\\    
\midrule
\cite{Aziz16})	& Lesion classification & US & DBN & DBN learns features from temporal US to classify prostate lesions benign/malignant\\
\cite{Shah16} 	& CBIR & MRI & CNN & Features from pre-trained CNN combined with features from hashing forest \\  
\cite{Zhu17} 	& Lesion classification & MRI & SAE & Learns features from multiple modalities, hierarchical random forest for classification \\
\midrule
{\em Bladder} &&&&\\    
\midrule
\cite{Cha16b} & Segmentation & CT & CNN & CNN patch classification used as initialization for level set\\ 
\bottomrule    
    \end{tabular}}%
  \label{tab:abdomen}%
\end{table*}%

\subsection{Cardiac}

Deep learning has been applied to many aspects of cardiac image analysis; the literature is summarized in Table \ref{tab:cardiac}. MRI is the most researched modality and left ventricle segmentation the most common task, but the number of applications is highly diverse: segmentation, tracking, slice classification, image quality assessment, automated calcium scoring and coronary centerline tracking, and super-resolution. 

Most papers used simple 2D CNNs and analyzed the 3D and often 4D data slice by slice; the exception is \cite{Wolt16} where 3D CNNs were used. DBNs are used in four papers, but these all originated from the same author group. The DBNs are only used for feature extraction and are integrated in compound segmentation frameworks. Two papers are exceptional because they combined CNNs with RNNs: \cite{Poud16} introduced a recurrent connection within the U-net architecture to segment the left ventricle slice by slice and learn what information to remember from the previous slices when segmenting the next one. \cite{Kong16} used an architecture with a standard 2D CNN and an LSTM to perform temporal regression to identify specific frames and a cardiac sequence.  
Many papers use publicly available data. The largest challenge in this field was the 2015 Kaggle Data Science Bowl where the goal was to automatically measure end-systolic and end-diastolic volumes in cardiac MRI. 192 teams competed for \$200,000 in prize money and the top ranking teams all used deep learning, in particular fCNN or U-net segmentation schemes. 

\subsection{Abdomen}
Most papers on the abdomen aimed to localize and segment organs, mainly the liver, kidneys, bladder, and pancreas (Table \ref{tab:abdomen}). Two papers address liver tumor segmentation. The main modality is MRI for prostate analysis and CT for all other organs. The colon is the only area where various applications were addressed, but always in a straightforward manner: A CNN was used as a feature extractor and these features were used for classification.

It is interesting to note that in two segmentation challenges - SLIVER07 for liver and PROMISE12 for prostate - more traditional image analysis methods were dominant up until 2016. In PROMISE12, the current second and third in rank among the automatic methods used active appearance models. The algorithm from IMorphics was ranked first for almost five years (now ranked second). However, a 3D fCNN similar to U-net \citep{Lequ17} has recently taken the top position. This paper has an interesting approach where a sum-operation was used instead of the concatenation operation used in U-net, making it a hybrid between a ResNet and U-net architecture. Also in SLIVER07 - a 10-year-old liver segmentation challenge - CNNs have started to appear in 2016 at the top of the leaderboard, replacing previously dominant methods focused on shape and appearance modeling.

\subsection{Musculoskeletal}
\begin{table*}[tb!]
  \centering
  \caption{Overview of papers using deep learning for musculoskeletal image analysis.}
    \resizebox{\textwidth}{!}{\begin{tabular}{lllllp{6cm}}
    \toprule
    Reference & Modality & Application; remarks \\
\midrule

\cite{Pras13} 	& MRI 		& Knee cartilage segmentation using multi-stream CNNs \\
\cite{Chen15e} 	& CT 		& Vertebrae localization; joint learning of vertebrae appearance and dependency on neighbors using CNN\\
\cite{Roth15b} 	& CT 		& Sclerotic metastases detection; random 2D views are analyzed by CNN and aggregated\\
\cite{Shen15a} 	& CT 		& Vertebrae localization and segmentation; CNN for segmenting vertebrae and for center detection\\
\cite{Suza15} 	& MRI 		& Vertebrae localization, identification and segmentation of vertebrae; CNN used for initial localization\\
\cite{Yang15b} 	& MRI 		& Anatomical landmark detection; uses CNN for slice classification for presence of landmark \\
\cite{Anto16}	& X-ray 	& Osteoarthritis grading; pre-trained ImageNet CNN fine-tuned on knee X-rays \\
\cite{Cai16}	& CT, MRI 	& Vertebrae localization; RBM determines position, orientation and label of vertebrae\\
\cite{Gola16} 	& US 		& Hip dysplasia detection; CNN with adversarial component detects structures and performs measurements\\
\cite{Kore16}	& MRI 		&  Vertebral bodies segmentation; voxel probabilities obtained with a 3D CNN are input to deformable model\\
\cite{Jama16} 	& MRI 		& Automatic spine scoring; VGG-19 CNN analyzes vertebral discs and finds lesion hotspots\\
\cite{Miao16a} 	& X-ray 	& Total Knee Arthroplasty kinematics by real-time 2D/3D registration using CNN\\
\cite{Roth16b} 	& CT 		& Posterior-element fractures detection; CNN for 2.5D patch-based analysis\\
\cite{Ster16} 	& MRI 		& Hand age estimation; 2D regression CNN analyzes 13 bones\\
\cite{Fors17}	& MRI 		& Vertebrae detection and labeling; outputs of two CNNs are input to graphical model\\
\cite{Spam17}	& X-ray 	& Skeletal bone age assessment; comparison among several deep learning approaches for the task at hand\\

\bottomrule    
    \end{tabular}}%
  \label{tab:musculo}%
\end{table*}%

Musculoskeletal images have also been analyzed by deep learning algorithms for segmentation and identification of bone, joint, and associated soft tissue abnormalities in diverse imaging modalities. The works are summarized in Table \ref{tab:musculo}. 

A surprising number of complete applications with promising results are available; one that stands out is \cite{Jama16} who trained their system with 12K discs and claimed near-human performances across four different radiological scoring tasks.

\subsection{Other}

\label{sec:applications_various}
\begin{table*}[tb!]
  \centering
  \caption{Overview of papers using a single deep learning approach for different tasks. DQN = Deep Q-Network}
    \resizebox{\textwidth}{!}{\begin{tabular}{lllllp{6cm}}
    \toprule
    Reference & Task & Modality & Method & Remarks \\
    \midrule
       
\cite{Shin13} 	& Heart, kidney, liver segmentation 	& MRI 			& SAE 		& SAE to learn temporal/spatial features on 2D + time DCE-MRI\\
\cite{Roth15e} 	& 2D slice classification 				& CT 			& CNN 		& Automatically classifying slices in 5 anatomical regions\\
\cite{Shin15} 	& 2D key image labeling 				& CT, MRI		& CNN 		& Text and 2D image analysis on a diverse set of 780 thousand images\\
\cite{Chen16} 	&	Various detection tasks 			& US, CT 		& AE, CNN 	& Detection of breast lesions in US and pulmonary nodules in CT \\
\cite{Ghes16a} 	& Landmark detection 					& US, CT, MRI 	& CNN, DQN 	& Reinforcement learning with CNN features, cardiac MR/US, head\&neck CT\\
\cite{Liu16a} 	& Image retrieval 						& X-ray 		& CNN 		& Combines CNN feature with Radon transform, evaluated on IRMA database \\
\cite{Merk16} 	& Vascular network segmentation 		& CT, MRI 		& CNN 		& Framework to find various vascular networks\\ 
\cite{Moes16a} 	& Various segmentation tasks 			& MRI, CT 		& CNN 		& Single architecture to segment 6 brain tissues, pectoral muscle \& coronaries\\
\cite{Roth16} 	& Various detection tasks 				& CT 			& CNN 		& Multi-stream CNN to detect sclerotic lesions, lymph nodes and polyps\\
\cite{Shin16} 	& Abnormality detection 				& CT 			& CNN 		& Compares architectures for detecting interstitial disease and lymph nodes\\
\cite{Tajb16} 	& Abnormality detection 				& CT, US 		& CNN 		& Compares pre-trained with fully trained networks for three detection tasks\\
\cite{Wang16b} 	& 2D key image labeling 				& CT, MRI		& CNN 		& Text concept clustering, related to \cite{Shin15}\\
\cite{Yan16} 	& 2D slice classification 				& CT	 		& CNN 		& Automatically classifying CT slices in 12 anatomical regions\\
\cite{Zhou16} 	& Thorax-abdomen segmentation 			& CT 			& CNN 		& 21 structures are segmented with 3 orthogonal 2D fCNNs and majority voting\\
\bottomrule    
    \end{tabular}}%
  \label{tab:various}%
\end{table*}%
\begin{table*}[tb!]
  \centering
  \caption{Overview of papers using deep learning for various image analysis tasks.}
    \resizebox{\textwidth}{!}{\begin{tabular}{lllllp{6cm}}
    \toprule
    Reference & Task & Modality & Method & Remarks \\
    \midrule

{\em Fetal imaging}&&&&\\
\midrule
\cite{Chen15b} 	& Frame labeling 						& US & CNN & Locates abdominal plane from fetal ultrasound videos\\
\cite{Chen15d} 	& Frame labeling 						& US & RNN & Same task as \cite{Chen15b}, now using RNNs\\
\cite{Baum16} 	& Frame labeling 						& US & CNN & Labeling 12 standard frames in 1003 mid pregnancy fetal US videos\\
\cite{Gao16a} 	& Frame labeling 						& US & CNN & 4 class frame classification using transfer learning with pre-trained networks \\
\cite{Kuma16} 	& Frame labeling 						& US & CNN & 12 standard anatomical planes, CNN extracts features for support vector machine\\
\cite{Rajc16} 	& Segmentation with non expert labels 	& MRI & CNN & Crowd-sourcing annotation efforts to segment brain structures \\ 
\cite{Rajc16a} 	& Segmentation given bounding box 		& MRI & CNN & CNN and CRF for segmentation of structures \\ 
\cite{Ravi16} 	& Quantification 						& US & CNN & Hybrid system using CNN and texture features to find abdominal circumference \\
\cite{Yu16} 	& Left ventricle segmentation 			& US & CNN &  Frame-by-frame segmentation by dynamically fine-tuning CNN to the latest frame\\
\midrule
{\em Dermatology}&&&&\\
\midrule
\cite{Code15} & \multicolumn{2}{l}{Melanoma detection in dermoscopic images}& CNN& Features from pre-trained CNN combined with other features \\
\cite{Demy16} & \multicolumn{2}{l}{Pattern identification in dermoscopic images} & CNN & Comparison to simpler networks and simple machine learning\\
\cite{Kawa16} & \multicolumn{2}{l}{5 and 10-class classification photographic images} & CNN & Pre-trained CNN for feature extraction at two image resolutions\\
\cite{Kawa16a} & \multicolumn{2}{l}{10-class classification photographic images}& CNN & Extending \cite{Kawa16} now training multi-resolution CNN end-to-end\\
\cite{Lequ16} & \multicolumn{2}{l}{Melanoma detection in dermoscopic images} & CNN & Deep residual networks for lesion segmentation and classification, winner ISIC16\\
\cite{Mene16} & \multicolumn{2}{l}{Classification of dermoscopic images} & CNN& Various pre-training and fine-tuning strategies are compared\\
\cite{Este17}  & \multicolumn{2}{l}{Classification of photographic and dermoscopic images} & CNN & Inception CNN trained on 129k images; compares favorably to 29 dermatologists\\

\midrule
{\em Lymph nodes}&&&&\\
\midrule
\cite{Roth14b} & Lymph node detection & CT & CNN & Introduces multi-stream framework of 2D CNNs with orthogonal patches\\
\cite{Barb16} & Lymph node detection & CT & CNN& Compares effect of different loss functions\\
\cite{Nogu16} & Lymph node detection & CT & CNN & 2 fCNNs, for inside and for contour of lymph nodes, are combined in a CRF\\

\midrule
{\em Other}&&&&\\
\midrule
\cite{Wang15d} & Wound segmentation & photographs & CNN & Additional detection of infection risk and healing progress \\
\cite{Ypsi15} & Chemotherapy response prediction & PET & CNN & CNN outperforms classical radiomics features in patients with esophageal cancer\\
\cite{Zhen15a} & Carotid artery bifurcation detection & CT & CNN & Two stage detection process, CNNs combined with Haar features\\
\cite{Alan16} &	Placenta segmentation & MRI & CNN & 3D multi-stream CNN with extension for motion correction \\
\cite{Frit16} & Head\&Neck tumor segmentation & CT & CNN & 3 orthogonal patches in 2D CNNs, combined with other features\\
\cite{Jaum16} &	Tongue contour extraction & US & RBM & Analysis of tongue motion during speech, combines auto-encoders with RBMs \\
\cite{Paye16} & Hand landmark detection  & X-ray & CNN & Various architectures are compared \\
\cite{Quin16} & Disease detection & microscopy & CNN & Smartphone mounted on microscope detects malaria, tuberculosis \& parasite eggs\\
\cite{Smis16} & Vessel detection and segmentation & US & CNN & Femoral and carotid vessels analyzed with standard fCNN \\
\cite{Twin16} & Task recognition in laparoscopy & Videos & CNN & Fine-tuned AlexNet applied to video frames \\
\cite{Xu16f} & Cervical dysplasia & cervigrams & CNN & Fine-tuned pre-trained network with added non-imaging features \\
\cite{Xue16} & Esophageal microvessel classification & Microscopy & CNN & Simple CNN used for feature extraction\\
\cite{Zhan16a} & Image reconstruction & CT & CNN & Reconstructing from limited angle measurements, reducing reconstruction artefacts \\
\cite{Leka17} & Carotid plaque classification & US & CNN & Simple CNN for characterization of carotid plaque composition in ultrasound\\
\cite{Ma17} & Thyroid nodule detection& US& CNN& CNN and standard features combines for 2D US analysis\\
\bottomrule    
    \end{tabular}}%
  \label{tab:other}%
\end{table*}%

This final section lists papers that address multiple applications (Table \ref{tab:various}) and a variety of other applications (Table \ref{tab:other}). 

It is remarkable that one single architecture or approach based on deep learning can be applied without modifications to different tasks; this illustrates the versatility of deep learning and its general applicability. In some works, pre-trained architectures are used, sometimes trained with images from a completely different domain. Several authors analyze the effect of fine-tuning a network by training it with a small data set of images from the intended application domain. Combining features extracted by a CNN with `traditional' features is also commonly seen.

From Table \ref{tab:other}, the large number of papers that address obstetric applications stand out. Most papers address the groundwork, such as selecting an appropriate frame from an US stream. More work on automated measurements with deep learning in these US sequences is likely to follow. 

The second area where CNNs are rapidly improving the state of the art is dermoscopic image analysis. For a long time, diagnosing skin cancer from photographs was considered very difficult and out of reach for computers. Many studies focused only on images obtained with specialized cameras, and recent systems based on deep networks produced promising results. A recent work by \cite{Este17} demonstrated excellent results with training a recent standard architecture (Google's Inception v3) on a data set of both dermoscopic and standard photographic images. This data set was two orders of magnitude larger than what was used in literature before. In a thorough evaluation, the proposed system performed on par with 30 board certified dermatologists. 

\color{black}
\section{Discussion}
\subsection*{Overview}
From the 308 papers reviewed in this survey, it is evident that deep learning has pervaded every aspect of medical image analysis. This has happened extremely quickly: the vast majority of contributions, 242 papers, were published in 2016 or the first month of 2017. A large diversity of deep architectures are covered. The earliest studies used pre-trained CNNs as feature extractors. The fact that these pre-trained networks could simply be downloaded and directly applied to any medical image facilitated their use. Moreover, in this approach already existing systems based on handcrafted features could simply be extended. In the last two years, however, we have seen that end-to-end trained CNNs have become the preferred approach for medical imaging interpretation (see Figure \ref{fig:statistics}). Such CNNs are often integrated into existing image analysis pipelines and replace traditional handcrafted machine learning methods. This is the approach followed by the largest group of papers in this survey and we can confidently state that this is the current standard practice.

\subsection*{Key aspects of successful deep learning methods}
After reviewing so many papers one would expect to be able to distill the perfect deep learning method and architecture for each individual task and application area. Although convolutional neural networks (and derivatives) are now clearly the top performers in most medical image analysis competitions, one striking conclusion we can draw is that the exact architecture is not the most important determinant in getting a good solution. We have seen, for example in challenges like the Kaggle Diabetic Retinopathy Challenge, that many researchers use the exact same architectures, the same type of networks, but have widely varying results. A key aspect that is often overlooked is that expert knowledge about the task to be solved can provide advantages that go beyond adding more layers to a CNN. Groups and researchers that obtain good performance when applying deep learning algorithms often differentiate themselves in aspects outside of the deep network, like novel data preprocessing or augmentation techniques. An example is that the best performing method in the CAMELYON16-challenge improved significantly (AUC from 0.92 to 0.99) by adding a stain normalization pre-processing step to improve generalization without changing the CNN. Other papers focus on data augmentation strategies to make networks more robust, and they report that these strategies are essential to obtain good performance. An example is the elastic deformations that were applied in the original U-Net paper \citep{Ronn15}.

Augmentation and pre-processing are, of course, not the only key contributors to good solutions. Several researchers have shown that designing architectures incorporating unique task-specific properties can obtain better results than straightforward CNNs. Two examples which we encountered several times are multi-view and multi-scale networks. Other, often underestimated, parts of network design are the network input size and receptive field (i.e.~the area in input space that contributes to a single output unit). Input sizes should be selected considering for example the required resolution and context to solve a problem. One might increase the size of the patch to obtain more context, but without changing the receptive field of the network this might not be beneficial. As a standard sanity check researchers could perform the same task themselves via visual assessment of the network input. If they, or domain experts, cannot achieve good performance, the chance that you need to modify your network input or architecture is high.

The last aspect we want to touch on is model hyper-parameter optimization (e.g.~learning rate, dropout rate), which can help squeeze out extra performance from a network. We believe this is of secondary importance with respect to performance to the previously discussed topics and training data quality. Disappointingly, no clear recipe can be given to obtain the best set of hyper-parameters as it is a highly empirical exercise. Most researchers fall back to an intuition-based random search \citep{Berg12d}, which often seems to work well enough. Some basic tips have been covered before by \cite{Beng12}. Researchers have also looked at Bayesian methods for hyper-parameter optimization \citep{Snoe12}, but this has not been applied in medical image analysis as far as we are aware of. 

\subsection*{Unique challenges in medical image analysis}
It is clear that applying deep learning algorithms to medical image analysis presents several unique challenges. The lack of large training data sets is often mentioned as an obstacle. However, this notion is only partially correct. The use of PACS systems in radiology has been routine in most western hospitals for at least a decade and these are filled with millions of images. There are few other domains where this magnitude of imaging data, acquired for specific purposes, are digitally available in well-structured archives. PACS-like systems are not as broadly used for other specialties in medicine, like ophthalmology and pathology, but this is changing as imaging becomes more prevalent across disciplines. We are also seeing that increasingly large public data sets are made available: \cite{Este17} used 18 public data sets and more than $10^5$ training images; in the Kaggle diabetic retinopathy competition a similar number of retinal images were released; and several chest x-ray studies used more than $10^4$ images. 

The main challenge is thus not the availability of image data itself, but the acquisition of relevant annotations/labeling for these images. Traditionally PACS systems store free-text reports by radiologists describing their findings. Turning these reports into accurate annotations or structured labels in an automated manner requires sophisticated text-mining methods, which is an important field of study in itself where deep learning is also widely used nowadays. With the introduction of structured reporting into several areas of medicine, extracting labels to data is expected to become easier in the future. For example, there are already papers appearing which directly leverage BI-RADS categorizations by radiologist to train deep networks \citep{Kisi16} or semantic descriptions in analyzing optical coherence tomography images \citep{Schl15}. We expect the amount of research in optimally leveraging free-text and structured reports for network training to increase in the near future.

Given the complexity of leveraging free-text reports from PACS or similar systems to train algorithms, generally researchers request domain experts (e.g.~radiologist, pathologists) to make task-specific annotations for the image data. Labeling a sufficiently large dataset can take a significant amount of time, and this is problematic. For example, to train deep learning systems for segmentation in radiology often 3D, slice-by-slice annotations need to be made and this is very time consuming. Thus, learning efficiently from limited data is an important area of research in medical image analysis. A recent paper focused on training a deep learning segmentation system for 3D segmentation using only sparse 2D segmentations \citep{Cice16}. Multiple-instance or active learning approaches might also offer benefit in some cases, and have recently been pursued in the context of deep learning \citep{Yan16}. One can also consider leveraging non-expert labels via crowd-sourcing \citep{Rajc16}. Other potential solutions can be found within the medical field itself; in histopathology one can sometimes use specific immunohistochemical stains to highlight regions of interest, reducing the need for expert experience \citep{Turk16}.

Even when data is annotated by domain expert, label noise can be a significant limiting factor in developing algorithms, whereas in computer vision the noise in the labeling of images is typically relatively low. To give an example, a widely used dataset for evaluating image analysis algorithms to detect nodules in lung CT is the LIDC-IDRI dataset \citep{Arma11}. In this dataset pulmonary nodules were annotated by four radiologists independently. Subsequently the readers reviewed each others annotations but no consensus was forced. It turned out that the number of  nodules they did not unanimously agreed on to be a nodule, was three times larger than the number they did fully agree on. Training a deep learning system on such data requires careful consideration of how to deal with noise and uncertainty in the reference standard. One could think of solutions like incorporating labeling uncertainty directly in the loss function, but this is still an open challenge.

In medical imaging often classification or segmentation is presented as a binary task: normal versus abnormal, object versus background. However, this is often a gross simplification as both classes can be highly heterogeneous. For example, the normal category often consists of completely normal tissue but also several categories of benign findings, which can be rare, and may occasionally include a wide variety of imaging artifacts. This often leads to systems that are extremely good at excluding the most common normal subclasses, but fail miserably on several rare ones. A straightforward solution would be to turn the deep learning system in a multi-class system by providing it with detailed annotations of all possible subclasses. Obviously this again compounds the issue of limited availability of expert time for annotating and is therefore often simply not feasible. Some researchers have specifically looked into tackling this imbalance by incorporating intelligence in the training process itself, by applying selective sampling \citep{Grin16b} or hard negative mining \citep{Wang16a}. However, such strategies typically fail when there is substantial noise in the reference standard.  Additional methods for dealing with within-class heterogeneity would be highly welcome.

Another data-related challenge is class imbalance. In medical imaging, images for the abnormal class might be challenging to find, depending on the task at hand. As an example, the implementation of breast cancer screening programs has resulted in vast databases of mammograms that have been established at many locations world-wide. However, the majority of these images are normal and do not contain any suspicious lesions. When a mammogram does contain a suspicious lesion this is often not cancerous, and even most cancerous lesions will not lead to the death of a patient. Designing deep learning systems that are adept at handling this class imbalance is another important area of research. A typical strategy we encountered in current literature is the application of specific data augmentation algorithms to just the underrepresented class, for example scaling and rotation transforms to generate new lesions. \cite{Pere16} performed a thorough evaluation of data augmentation strategies for brain lesion segmentation to combat class imbalance.

In medical image analysis useful information is not just contained within the images themselves. Physicians often leverage a wealth of data on patient history, age, demographics and others to arrive at better decisions. Some authors have already investigated combining this information into deep learning networks in a straightforward manner \citep{Kooi17}. However, as these authors note, the improvements that were obtained were not as large as expected. One of the challenges is to balance the number of imaging features in the deep learning network (typically thousands) with the number of clinical features (typically only a handful) to prevent the clinical features from being drowned out. Physicians often also need to use anatomical information to come to an accurate diagnosis. However, many deep learning systems in medical imaging are still based on patch classification, where the anatomical location of the patch is often unknown to network. One solution would be to feed the entire image to the deep network and use a different type of evaluation to drive learning, as was done by, for example, \cite{Mill16}, who designed a loss function based on the Dice coefficient. This also takes advantage of the fact that medical images are often acquired using a relatively static protocol, where the anatomy is always roughly in the same position and at the same scale. However, as mentioned above, if the receptive field of the network is small feeding in the entire image offers no benefit. Furthermore, feeding full images to the network is not always feasible due to, for example, memory constraints. In some cases this might be solved in the near future due to advances in GPU technology, but in others, for example digital pathology with its gigapixel-sized images, other strategies have to be invented.

\subsection*{Outlook}
Although most of the challenges mentioned above have not been adequately tackled yet, several high-profile successes of deep learning in medical imaging have been reported, such as the work by \cite{Este17} and \cite{Guls16} in the fields of dermatology and ophthalmology. Both papers show that it is possible to outperform medical experts in certain tasks using deep learning for image classification. However, we feel it is important to put these papers into context relative to medical image analysis in general, as most tasks can by no means be considered 'solved'. One aspect to consider is that both \cite{Este17} and \cite{Guls16} focus on small 2D color image classification, which is relatively similar to the tasks that have been tackled in computer vision (e.g. ImageNet). This allows them to take advantage of well-explored network architectures like ResNet and VGG-Net which have shown to have excellent results in these tasks. However, there is no guarantee that these architectures are optimal in for example regressions/detection tasks. It also allowed the authors to use networks that were pre-trained on a very well-labeled dataset of millions of natural images, which helps combat the lack of similarly large, labeled medical datasets. In contrast, in most medical imaging tasks 3D gray-scale or multi-channel images are used for which pre-trained networks or architectures don’t exist. In addition this data typically has very specific challenges, like anisotropic voxel sizes, small registration errors between varying channels (e.g. in multi-parametric MRI) or varying intensity ranges. Although many tasks in medical image analysis can be postulated as a classification problem, this might not always be the optimal strategy as it typically requires some form of post-processing with non-deep learning methods (e.g. counting, segmentation or regression tasks). An interesting example is the paper by \cite{Siri16}, which details a method directly predicting the center locations of nuclei and shows that this outperforms classification-based center localization. Nonetheless, the papers by \cite{Este17} and \cite{Guls16} do show what ideally is possible with deep learning methods that are well-engineered for specific medical image analysis tasks.

Looking at current trends in the machine learning community with respect to deep learning, we identify a key area which can be highly relevant for medical imaging and is receiving (renewed) interest: unsupervised learning. The renaissance of neural networks started around 2006 with the popularization of greedy layer-wise pre-training of neural networks in an unsupervised manner. This was quickly superseded by fully supervised methods which became the standard after the success of AlexNet during the ImageNet competition of 2012, and most papers in this survey follow a supervised approach. However, interest in unsupervised training strategies has remained and recently has regained traction.

Unsupervised methods are attractive as they allow (initial) network training with the wealth of unlabeled data available in the world. Another reason to assume that unsupervised methods will still have a significant role to play is the analogue to human learning, which seems to be much more data efficient and also happens to some extent in an unsupervised manner; we can learn to recognize objects and structures without knowing the specific label. We only need very limited supervision to categorize these recognized objects into classes. Two novel unsupervised strategies which we expect to have an impact in medical imaging are variational auto-encoders (VAEs), introduced by \cite{King13} and generative adversarial networks (GANs), introduced by \cite{Good14}. The former merges variational Bayesian graphical models with neural networks as encoders/decoders. The latter uses two competing convolutional neural networks where one is generating artificial data samples and the other is discriminating artificial from real samples. Both have stochastic components and are generative networks. Most importantly, they can be trained end-to-end and learn representative features in a completely unsupervised manner. As we discussed in previous paragraphs, obtaining large amounts of unlabeled medical data is generally much easier than labeled data and unsupervised methods like VAEs and GANs could optimally leverage this wealth of information.

Finally, deep learning methods have often been described as `black boxes'. Especially in medicine, where accountability is important and can have serious legal consequences, it is often not enough to have a good prediction system. This system also has to be able to articulate itself in a certain way. Several strategies have been developed to understand what intermediate layers of convolutional networks are responding to, for example deconvolution networks \citep{Zeil14}, guided back-propagation \citep{Spri14} or deep Taylor composition \citep{Mont17}. Other researchers have tied prediction to textual representations of the image (i.e.~captioning) \citep{Karp15}, which is another useful avenue to understand what a network is perceiving. Last, some groups have tried to combine Bayesian statistics with deep networks to obtain true network uncertainty estimates \cite{Kend17}. This would allow physicians to assess when the network is giving unreliable predictions. Leveraging these techniques in the application of deep learning methods to medical image analysis could accelerate acceptance of deep learning applications among clinicians, and among patients. We also foresee deep learning approaches will be used for related tasks in medical imaging, mostly unexplored, such as image reconstruction \citep{Wang16d}. Deep learning will thus not only have a great impact in medical image analysis, but in medical imaging as a whole.
\color{black}

\section*{Acknowledgments}
The authors would like to thank members of the Diagnostic Image Analysis Group for discussions and suggestions. This research was funded by grants KUN 2012-5577, KUN 2014-7032, and KUN 2015-7970 of the Dutch Cancer Society.

\section*{Appendix A: Literature selection}
\color{black}
PubMed was searched for papers containing "convolutional" OR "deep learning" in any field. We specifically did not include the term neural network here as this would result in an enormous amount of 'false positive' papers covering brain research. This search initially gave over 700 hits. ArXiv was searched for papers mentioning one of a set of terms related to medical imaging. The exact search string was: 'abs:((medical OR mri OR "magnetic resonance" OR CT OR "computed tomography" OR ultrasound OR pathology OR xray OR x-ray OR radiograph OR mammography OR fundus OR OCT) AND ("deep learning" OR convolutional OR cnn OR "neural network"))'. Conference proceedings for MICCAI (including workshops), SPIE, ISBI and EMBC were searched based on titles of papers. Again we looked for mentions of 'deep learning' or 'convolutional' or 'neural network'. We went over all these papers and excluded the ones that did not discuss medical imaging (e.g. applications to genetics, chemistry), only used handcrafted features in combination with neural networks, or only referenced deep learning as future work. When in doubt whether a paper should be included we read the abstract and when the exact methodology was still unclear we read the paper itself. We checked references in all selected papers iteratively and consulted colleagues to identify any papers which were missed by our initial search. When largely overlapping work had been reported in multiple publications, only the publication deemed most important was included. A typical example here was arXiv preprints that were subsequently published or conference contributions which were expanded and published in journals.
\color{black}
\section*{References}
\bibliographystyle{elsarticle-harv}
\bibliography{medlinestrings,dl}
\end{document}